\pdfoutput=1

\documentclass[11pt]{article}
\usepackage{paralist}
\usepackage[final]{acl}

\usepackage{times}
\usepackage{latexsym}
\usepackage{multirow}
\usepackage{multicol}
\usepackage{booktabs}
\usepackage[table]{xcolor}
\usepackage{graphicx}
\usepackage{arydshln}
\usepackage{makecell}
\usepackage{amssymb} 
\usepackage{subcaption} 
\usepackage{colortbl}
\usepackage{booktabs}
\usepackage{tabularx}
\usepackage{arydshln} 
\usepackage{amsmath}
\usepackage{float}
\usepackage{pifont}
\usepackage[most]{tcolorbox}

\newcommand{\cmark}{{\color{green!60!black}\ding{51}}}
\newcommand{\xmark}{{\color{red!70!black}\ding{55}}}

\newtcolorbox{promptbox}{
  enhanced,
  breakable,                            
  colback=gray!5,
  colframe=gray!60,
  boxrule=0.6pt,
  left=2mm, right=2mm, top=1mm, bottom=1mm,
  fontupper=\small,
  sharp corners,
  before skip=6pt, after skip=6pt,      
  attach boxed title to top center={yshift=-1mm},
  boxed title style={
    colback=gray!20,
    sharp corners,
    size=small,
    boxrule=0.5pt
  },
  title=\textcolor{black}{\textbf{Prompt Template}},
}

\definecolor{teal0}{RGB}{220,245,245}   
\definecolor{teal1}{RGB}{180,230,230}   
\definecolor{teal2}{RGB}{120,200,200}   
\definecolor{teal3}{RGB}{60,160,160}    
\definecolor{teal4}{RGB}{0,128,128}     

\usepackage[T1]{fontenc}

\usepackage[utf8]{inputenc}

\usepackage{microtype}

\usepackage{inconsolata}

\usepackage{graphicx}

\makeatletter
\renewcommand{\thanks}[1]{%
  \begingroup
    \renewcommand{\@makefnmark}{\textcolor{black}{\@thefnmark}}
    \footnotemark
    \protected@xdef\@thanks{\@thanks\noexpand\footnotetext{#1}}%
  \endgroup
}
\makeatother

\title{Revisiting Metric Reliability for Fine-grained Evaluation of Machine Translation and Summarization in Indian Languages}

\author{
 \textbf{Amir Hossein Yari\textsuperscript{1,4}}\quad
 \textbf{Kalmit Kulkarni\textsuperscript{2}} \quad
 \textbf{Ahmad Raza Khan\textsuperscript{3\thanks{This author contributed solely by coordinating annotators.}}} \quad
 \textbf{Fajri Koto\textsuperscript{4}}
\\
\\
 Sharif University of Technology\textsuperscript{1}\quad
 Vellore Institute of Technology\textsuperscript{2}\\
 IIT Kharagpur\textsuperscript{3}\quad
 Mohamed bin Zayed University of Artificial Intelligence\textsuperscript{4}
\\
 {
   \texttt{\small ahyari2002@gmail.com},  \texttt{\small fajri.koto@mbzuai.ac.ae}
 }
}

\begin{document}
\maketitle
\begin{abstract}
While automatic metrics drive progress in Machine Translation (MT) and Text Summarization (TS), existing metrics have been developed and validated almost exclusively for English and other high-resource languages. This narrow focus leaves Indian languages—spoken by over 1.5 billion people—largely overlooked, casting doubt on the universality of current evaluation practices. To address this gap, we introduce \texttt{\textbf{ITEM}}, a large-scale benchmark that systematically evaluates the alignment of 29 automatic metrics with human judgments across six major Indian languages, enriched with fine-grained annotations. Our extensive evaluation—covering agreement with human judgments, sensitivity to outliers, language-specific reliability, inter-metric correlations, and resilience to controlled perturbations—reveals four central findings: (1) LLM-based evaluators show the strongest alignment with human judgments at both segment and system levels; (2) outliers exert a significant impact on metric-human agreement; (3) In TS, metrics are more effective at capturing content fidelity, whereas in MT, they better reflect fluency; and (4) Metrics differ in their robustness and sensitivity when subjected to diverse perturbations. Collectively, these findings offer critical guidance for advancing metric design and evaluation in Indian languages.\footnote{This dataset is publicly released under a Creative Commons license at \url{https://huggingface.co/datasets/AmirHossein2002/ITEM}}%

\end{abstract}

\section{Introduction}

Effective deployment of Machine Translation (MT) and Text Summarization (TS) systems hinges on reliable evaluation. While human judgments remain the gold standard, their high cost, slow turnaround, and inherent subjectivity limit scalability and complicate large-scale or real-time assessment. Consequently, automatic metrics serve as practical proxies. Traditional surface-based metrics (e.g., BLEU \citep{papineni-etal-2002-bleu}, ROUGE \citep{lin-2004-rouge}) and modern embedding- or pretrained-based scorers (e.g., BERTScore \citep{zhang2020bertscoreevaluatingtextgeneration}, COMET \citep{rei-etal-2020-comet}) offer more efficient alternatives. However, these metrics were originally developed and validated on datasets in high-resource languages such as English and German, raising questions about their reliability and generalizability to other languages, particularly low-resource Indian languages. At the same time, most research in MT and TS continues to rely on these metrics without empirical evidence of their efficacy \cite{datta-etal-2023-mildsum,urlana-etal-2023-pmindiasum,kumar-etal-2025-cosmmic}.

Despite the rich linguistic diversity and widespread use of Indian languages, prior research has largely overlooked them in the evaluation of MT and TS systems, relying on a narrow set of metrics and lacking fine-grained analysis. Existing efforts, such as \citet{sai-b-etal-2023-indicmt}, have assessed metric reliability only for MT and primarily at the overall quality level, leaving other tasks and evaluation dimensions unexplored. In this work, we broaden the scope to both MT and TS, introducing four and two fine-grained evaluation criteria, respectively, across six major Indian languages. 

Another limitation of prior work on metric evaluation lies in the limited consideration of robustness, which we address in this paper along two dimensions. First, previous studies often ignore the impact of outliers. As noted by \citet{mathur-etal-2020-tangled}, correlation measures such as Pearson’s can produce spurious correlations when outliers are present, yet works in Indian languages like \citet{sai-b-etal-2023-indicmt} overlook this issue. Second, most recent evaluations rely on high-quality translations or summaries, which weakens robustness analysis by offering only shallow signals—such as token overlap or embedding similarity—that do not always align with human perception. In the context of Indian languages, assessing metric robustness is particularly important, especially under controlled perturbations such as paraphrasing, negation, syntactic reordering, and lexical substitutions (e.g., synonyms or antonyms). Given their flexible word order and a wide range of morphological complexity—from moderate to highly inflectional—Indian languages exemplify the challenges of robust metric evaluation \cite{tandon-sharma-2017-unity}.



Our contributions can be summarized as follows:
\begin{compactitem}
\item We introduce \texttt{\textbf{ITEM}} (\textbf{I}ndian \textbf{T}ext \textbf{E}valuation \textbf{M}etrics Testbed), a benchmark for assessing the alignment between automatic metrics and human judgments in MT and TS across six major Indian languages: Hindi, Bengali, Marathi, and Gujarati (Indo-Aryan), and Tamil and Telugu (Dravidian).
\item We conduct a comprehensive evaluation of 29 automatic metrics to identify those that best align with human judgments at both segment and system levels.
\item We examine how effectively existing metrics capture diverse evaluation dimensions, including adequacy, fluency, coherence, and factuality.
\item We investigate the effect of outlier samples on metric–human correlations and evaluate metric robustness and sensitivity under controlled perturbations such as paraphrasing, negation, and syntactic reordering.
\end{compactitem}

\section{Related Work}


\paragraph{Evaluation of Machine Translation System} BLEU \cite{papineni-etal-2002-bleu}, ChrF++ \cite{popovic-2017-chrf}, COMET \cite{rei-etal-2020-comet}, and LLM-based scores \cite{yuan2021bartscoreevaluatinggeneratedtext,fu-etal-2024-gptscore} are among the most commonly used metrics for evaluating MT. However, these evaluations have generally been limited in scope, focusing on a narrow range of languages and metrics. Although some recent studies \cite{wu-etal-2024-evaluating-automatic, agrawal-etal-2024-automatic-metrics, freitag-etal-2024-llms} provide broader quality assessments, they entirely exclude Indian languages. Existing works that do include Indian languages \cite{sai-b-etal-2023-indicmt, singh-etal-2024-good} remain incomplete, as they report only overall scores, overlook multiple evaluation dimensions, and cover fewer languages than our study.

\paragraph{Evaluation of Text Summarization Systems}
The most commonly used metrics for evaluating TS are ROUGE \cite{lin-2004-rouge}, BERTScore \cite{zhang2020bertscoreevaluatingtextgeneration}, and LLM-based scores \cite{fu-etal-2024-gptscore}. In contrast, older metrics such as BLEU \cite{papineni-etal-2002-bleu} and METEOR \cite{banerjee-lavie-2005-meteor}, as well as newer ones like MoverScore \cite{zhao-etal-2019-moverscore}, are less frequently adopted. The scope of TS evaluation has also expanded beyond recall-based coverage to include additional dimensions such as focus, faithfulness, and inter-sentential coherence \cite{koto2022ffci,fabbri2021summevalreevaluatingsummarizationevaluation}.

Existing studies, however, have primarily focused on English, with only limited multilingual efforts \cite{koto-etal-2021-evaluating, clark-etal-2023-seahorse, han-etal-2024-rethinking} extending coverage to other languages. These works still exclude Indian languages and rely on a narrow set of traditional metrics and evaluation aspects.



\paragraph{Robustness Studies on Evaluation Systems} The robustness of evaluation metrics under perturbations has received limited attention. \citet{guo-etal-2024-appls} investigates controlled perturbations for English Plain Language Summarization.\footnote{A task of generating semantically faithful yet structurally simplified summaries to improve accessibility for readers with limited linguistic or domain expertise.} \citet{huang-baldwin-2023-robustness} evaluates MT metrics under word- and character-level attacks, yet excludes TS, Indian languages, and a broader range of perturbations. Other works \cite{alves-etal-2022-robust, chen-etal-2022-exploring} focus solely on MT, while \citep{patel-etal-2024-tweak} considers only English TS. None provides a comprehensive robustness assessment of both TS and MT metrics across Indian languages.

In contrast to prior work, we present a unified, high-resolution evaluation framework for both TS and MT in Indian languages. Our study goes beyond single overall scores by assessing multiple quality dimensions using 29 diverse metrics across varied configurations (including 48 ROUGE, 144 BERTScore, and 4 BLEU variants). We further evaluate robustness through controlled paraphrasing and perturbation tests, providing a benchmark-level, fine-grained analysis that surpasses the scope of existing studies.

\section{ITEM}
\label{sec:ITEM}
\subsection{Dataset Construction}
The \texttt{\textbf{ITEM}} benchmark was developed by assembling parallel datasets for summarization and translation in six widely spoken Indian languages: Hindi, Bengali, Tamil, Telugu, Marathi, and Gujarati. For summarization, we randomly sampled 150 article--summary pairs per language from the XLSum corpus \citep{hasan-etal-2021-xl}, which provides professionally annotated summaries of BBC news articles. For translation, we randomly selected 150 sentence--translation pairs per language from FLORES-200 \citep{nllbteam2022languageleftbehindscaling}, a widely used benchmark with human-verified translations. All translation outputs are annotated in the respective Indian languages.

\begin{figure*}[t]
  \begin{center}
    \includegraphics[width=\textwidth]{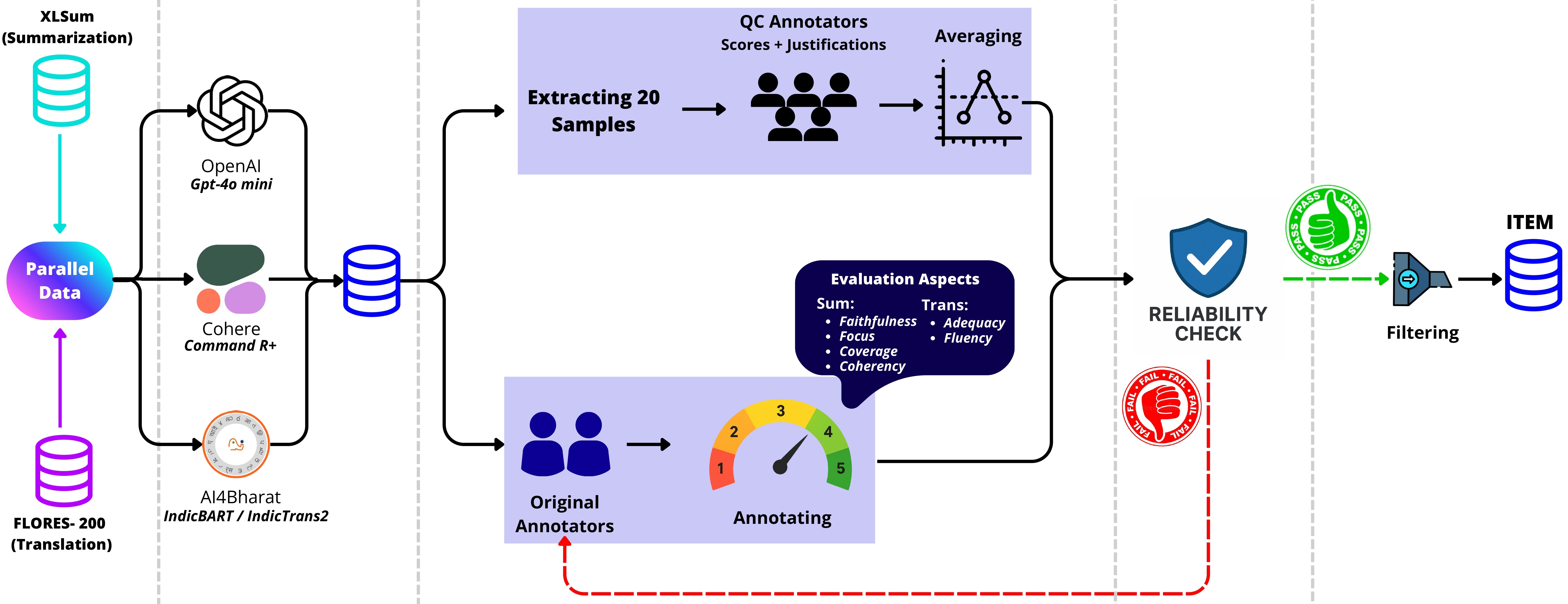}
  \end{center}
  \caption{End-to-End process of dataset creation.}
  \vspace{-0cm}
  \label{fig:annotation_pipeline}
\end{figure*}

To obtain a broad spectrum of machine-generated outputs, we applied three complementary models to the sampled data. The set comprises two state-of-the-art proprietary LLMs—Cohere Command R+ \citep{cohere_for_ai_2024} and GPT-4o mini \citep{openai2024gpt4ocard}—alongside IndicBART and IndicTrans2 from AI4Bharat \citep{dabre-etal-2022-indicbart, gala2023indictrans2highqualityaccessiblemachine}, open-source models designed with a focus on Indian languages. Please refer to Figure~\ref{fig:annotation_pipeline} for the complete annotation process \texttt{\textbf{ITEM}}.


\subsection{Evaluation Aspects} 
To account for the multifaceted nature of MT and TS quality, \texttt{\textbf{ITEM}} employs a set of carefully adopted quality indicators. For MT, evaluation centers on two fundamental dimensions \cite{koehn-monz-2006-manual}: (1) \textit{Adequacy}, which measures how accurately the translation conveys the meaning and intent of the source without omissions or distortions, and (2) \textit{Fluency}, which assesses the readability of the translated text in the target language, focusing on grammaticality, natural expression, and stylistic appropriateness.

For TS, a more nuanced four-dimensional framework is adopted to reflect the task’s inherent complexity \cite{koto2022ffci}. (1) \textit{Faithfulness} measures factual consistency, ensuring the summary avoids hallucinations or contradictions. (2) \textit{Focus} assesses the precision of the summary, ensuring only relevant information is included. (3) \textit{Coverage} examines how well the summary captures the key and salient ideas of the source. (4) \textit{Coherency} measures the logical continuity between sentences, ensuring that the summary maintains a seamless flow and remains readily understandable. 

\subsection{Quality Control} 

\begin{table}[t!]
\centering
\renewcommand{\arraystretch}{1} 
\setlength{\tabcolsep}{8pt}       
\resizebox{0.8\linewidth}{!}{%
\begin{tabular}{lcccc}
\toprule
\multirow{2}{*}{\textbf{Language}} & \multicolumn{2}{c}{\textbf{Worker 1}} & \multicolumn{2}{c}{\textbf{Worker 2}} \\
\cmidrule(lr){2-3} \cmidrule(lr){4-5}
                                   & \textbf{Sum.} & \textbf{Trans.} & \textbf{Sum.} & \textbf{Trans.} \\
\midrule
\rowcolor{gray!10} Hindi    & 81.3 & 92.5 & 86.3 & 85.0 \\
Bengali                        & 80.0 & 77.5 & 77.5 & 85.0 \\
\rowcolor{gray!10} Tamil    & 85.0 & 82.5 & 76.3 & 82.5 \\
Telugu                          & 92.5 & 87.5 & 87.5 & 87.5 \\
\rowcolor{gray!10} Marathi  & 76.3 & 77.5 & 82.5 & 87.5 \\
Gujarati                        & 87.5  & 87.5 & 91.3 & 90.0 \\
\bottomrule
\end{tabular}%
}
\vspace{-0.1cm}
\caption{Annotators quality (\%).}
\vspace{-0.2cm}
\label{tab:worker_quality}
\end{table}


Reliability of human annotations was a cornerstone in developing \texttt{\textbf{ITEM}}. To ensure consistency, we first established detailed annotation guidelines and recruited \textbf{two} native speakers per target language to evaluate the extracted samples. All annotators were qualified experts, each holding at least a high school diploma and proficient in both English and the respective Indian language. Annotators scored each sample against the predefined quality indicators using a five-point scale (1 = Poor, 5 = Excellent), with every level explicitly defined to minimize ambiguity and reduce subjective variance.

Beyond initial annotation, we implemented a structured quality control (QC) procedure. For each task and language, 20 representative samples\footnote{These 20 samples, encompassing both high- and low-quality data, were strategically selected to rigorously assess annotator sensitivity across the quality spectrum and strengthen QC reliability.
} were selected and assessed by five independent  expert annotators\footnote{All annotators in this study were aged 17–22, native speakers of their respective language, and proficient in English. Each worker is fairly compensated in accordance with an agreement, with payments aligned to India’s minimum wage.}, who provided both scores and written justifications. These QC annotations were carefully reviewed to verify alignment and reliability. Final reference scores for each of the 20 QC samples were obtained by averaging the ratings from all five annotators.

The reliability of our two main workers was then evaluated against these 20 QC samples. Specifically, we computed the proportion of annotations by each worker that deviated by at most one point from the QC reference scores across all evaluation aspects. This metric served as a robust and interpretable measure of annotator quality. The resulting quality scores are summarized in Table~\ref{tab:worker_quality}.

\subsection{Data Statistics}
The initial release of \texttt{\textbf{ITEM}} included 5,400 samples (2 tasks $\times$ 150 datapoints $\times$ 3 models $\times$ 6 languages). To enhance the benchmark's reliability, we filtered out samples with high annotator disagreement, resulting in 2,571 samples for summarization and 2,604 for translation, prioritizing annotation consistency over sheer volume. Table~\ref{tab:workers_correlation} reports inter-annotator correlations post-refinement, capturing alignment between annotators.

\begin{table}[ht!]
\centering
\renewcommand{\arraystretch}{1}
\setlength{\tabcolsep}{3pt}
\resizebox{0.9\columnwidth}{!}{%
\begin{tabular}{lcccccccc}
\toprule
\multirow{2}{*}{\textbf{Language}} & \multicolumn{5}{c}{\textbf{Summarization}} & \multicolumn{3}{c}{\textbf{Translation}} \\
\cmidrule(lr){2-6} \cmidrule(lr){7-9}
 & \textbf{Fa.} & \textbf{Fo.} & \textbf{Cov.} & \textbf{Coh.} & \textbf{Avg} & \textbf{Ad.} & \textbf{Fl.} & \textbf{Avg} \\
\midrule
\rowcolor{gray!10} Hindi      & 0.53 & 0.53 & 0.48 & 0.55 & 0.74 & 0.54 & 0.63 & 0.61 \\
Bengali                         & 0.8 & 0.82 & 0.84 & 0.4 & 0.93 & 0.54 & 0.47 & 0.64 \\
\rowcolor{gray!10} Tamil       & 0.45 & 0.57 & 0.68 & 0.71 & 0.81 & 0.65 & 0.61 & 0.72 \\
Telugu                           & 0.74 & 0.69 & 0.72 & 0.73 & 0.89 & 0.78 & 0.76 & 0.87 \\
\rowcolor{gray!10} Marathi     & 0.58 & 0.57 & 0.55 & 0.58 & 0.79 & 0.59 & 0.49 & 0.44 \\
Gujarati                         & 0.85 & 0.83 & 0.86 & 0.65 & 0.91 & 0.75 & 0.75 & 0.77 \\
\bottomrule
\end{tabular}%
}
\vspace{-0.1cm}
\caption{Inter-annotator Pearson correlations across evaluation aspects: Faithfulness (Fa), Focus (Fo), Coverage (Cov), Coherence (Coh), Adequacy (Ad), and Fluency (Fl).}
\vspace{-0cm}
\label{tab:workers_correlation}
\end{table}

Tables~\ref{tab:token_stats} and \ref{tab:machine_token_counts} present the average token counts for source texts alongside human references, as well as machine-generated summaries and translations, providing a concise overview of text length variations across different tasks and languages.

\begin{table}[t]
\centering
\renewcommand{\arraystretch}{1}
\setlength{\tabcolsep}{6pt}
\resizebox{0.9\linewidth}{!}{%
\begin{tabular}{lcccc}
\toprule
\textbf{Language} & \textbf{Article} & \textbf{Summary} & \textbf{English Text} & \textbf{Translation} \\
\midrule
\rowcolor{gray!10} Hindi     & 2,476 & 145 & 139 & 136 \\
Bengali   & 3,344 & 151 & 139 & 135 \\
\rowcolor{gray!10} Tamil     & 3,516 & 195 & 139 & 161 \\
Telugu    & 3,930 & 170 & 139 & 138 \\
\rowcolor{gray!10} Marathi   & 3,601 & 176 & 139 & 142 \\
Gujarati  & 3,259 & 136 & 139 & 133 \\
\bottomrule
\end{tabular}%
}
\vspace{-0.2cm}
\caption{Average token counts for source texts and human-generated summaries/translations.}
\vspace{-0.2cm}
\label{tab:token_stats}
\end{table}

\begin{table}[t]
\centering
\renewcommand{\arraystretch}{1} 
\setlength{\tabcolsep}{6pt}       
\resizebox{0.9\linewidth}{!}{%
\begin{tabular}{l c c c c c c}
\toprule
\multirow{2}{*}{\textbf{Language}}
  & \multicolumn{2}{c}{\textbf{Cohere}}
  & \multicolumn{2}{c}{\textbf{GPT}}
  & \multicolumn{2}{c}{\textbf{AI4Bharat}} \\
\cmidrule(lr){2-3} \cmidrule(lr){4-5} \cmidrule(lr){6-7}
 & \textbf{Sum.} & \textbf{Trans.}
 & \textbf{Sum.} & \textbf{Trans.}
 & \textbf{Sum.} & \textbf{Trans.} \\
\midrule
\rowcolor{gray!10} Hindi    & 239 & 140 & 268 & 136 & 194 & 142 \\
Bengali  & 138 & 135 & 248 & 136 & 228 & 136 \\
\rowcolor{gray!10} Tamil    & 120 & 162 & 245 & 157 & 245 & 158 \\
Telugu   &  82 & 143 & 222 & 138 & 228 & 137 \\
\rowcolor{gray!10} Marathi  & 149 & 138 & 241 & 137 & 231 & 139 \\
Gujarati &  84 & 140 & 223 & 132 & 183 & 136 \\
\bottomrule
\end{tabular}%
}
\vspace{-0.2cm}
\caption{Average token counts for machine-generated summaries/translations.}
\vspace{-0.2cm}
\label{tab:machine_token_counts}
\end{table}

\begin{figure*}[t!]
    \centering
    \begin{subfigure}{0.475\textwidth}
        \centering
        \includegraphics[width=\linewidth]{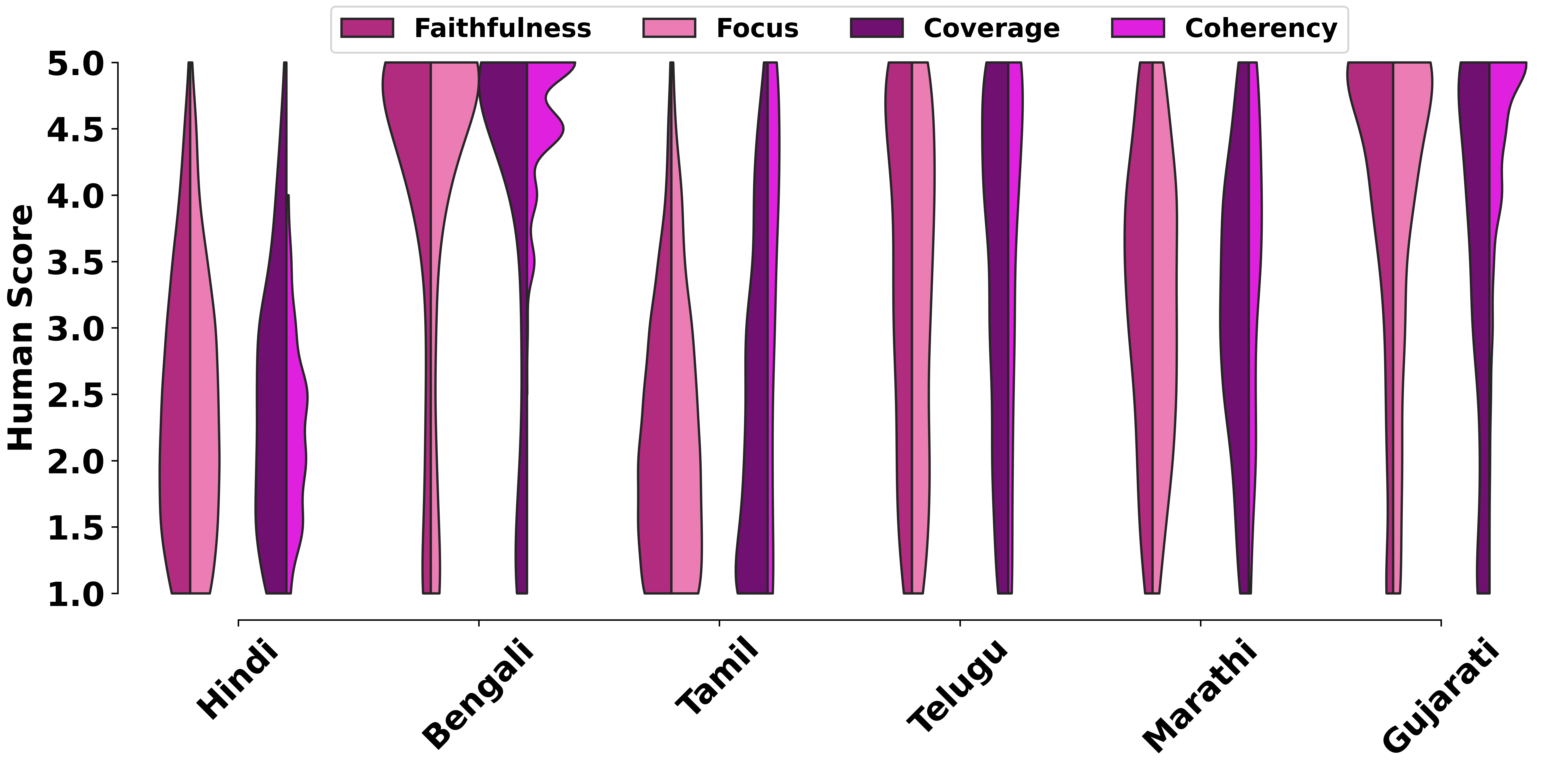}
        \caption{Text Summarization}
        \label{fig:summ_scores}
    \end{subfigure}
    \hfill
    \begin{subfigure}{0.475\textwidth}
        \centering
        \includegraphics[width=\linewidth]{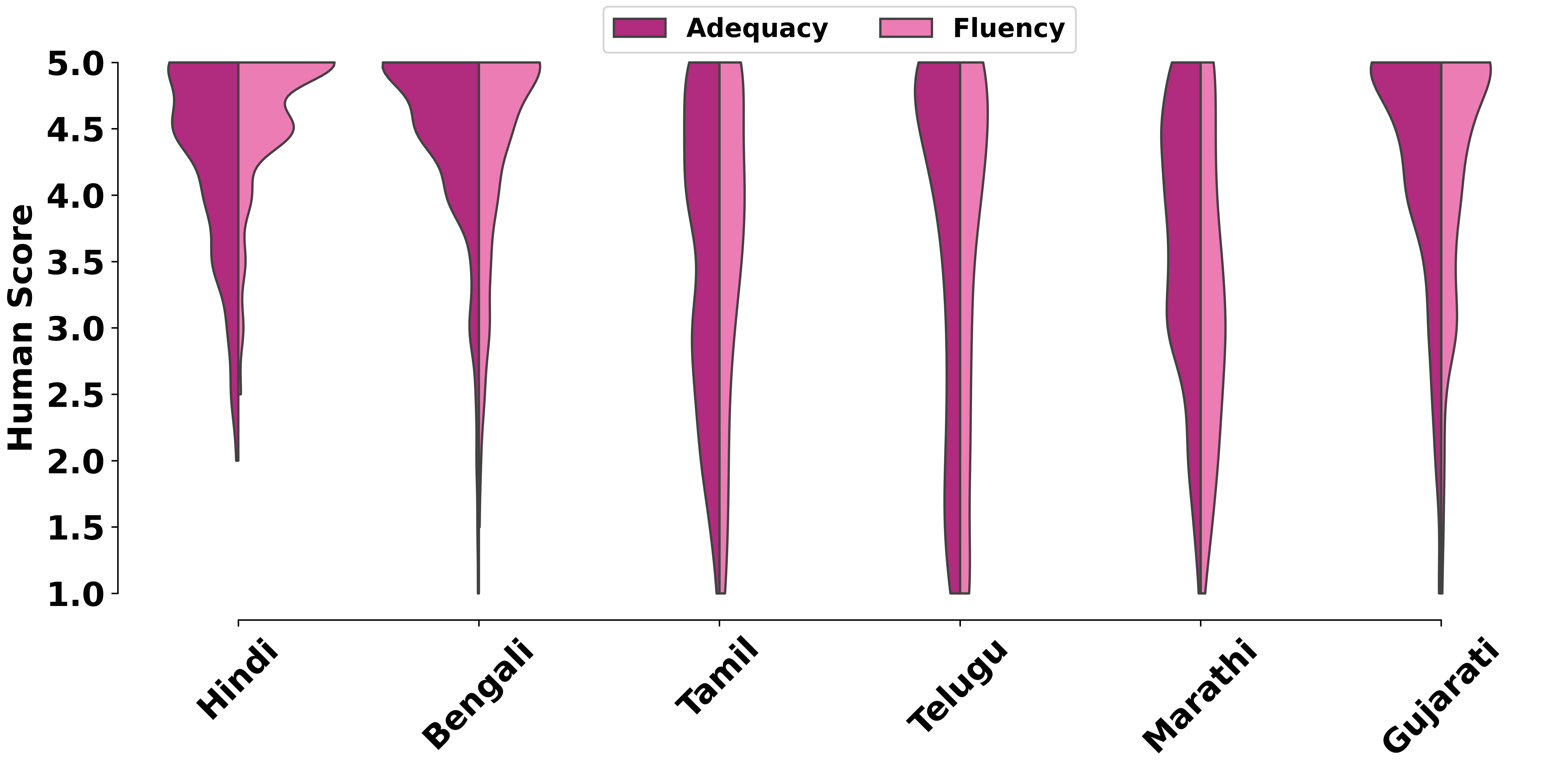}
        \caption{Machine Translation}
        \label{fig:trans_scores}
    \end{subfigure}
    \vspace{-0.15cm}
    \caption{Distribution of human evaluation scores across evaluation aspects.}
    \label{fig:human_scores}
\end{figure*}

We define human scores as the average of the ratings provided by two annotators. Figure~\ref{fig:human_scores} depicts the distribution of human evaluation scores. In TS, Bengali attains the highest quality ratings, while Hindi ranks lowest. Conversely, in MT, Hindi leads in quality, with Marathi at the bottom. A deeper examination of these scores reveals that GPT-4o mini consistently excels in TS across all languages, whereas IndicTrans2 dominates MT for most languages, with GPT-4o mini outperforming only in Hindi. 

\begin{figure}[t]
  \begin{center}
    \includegraphics[width=0.7\columnwidth]{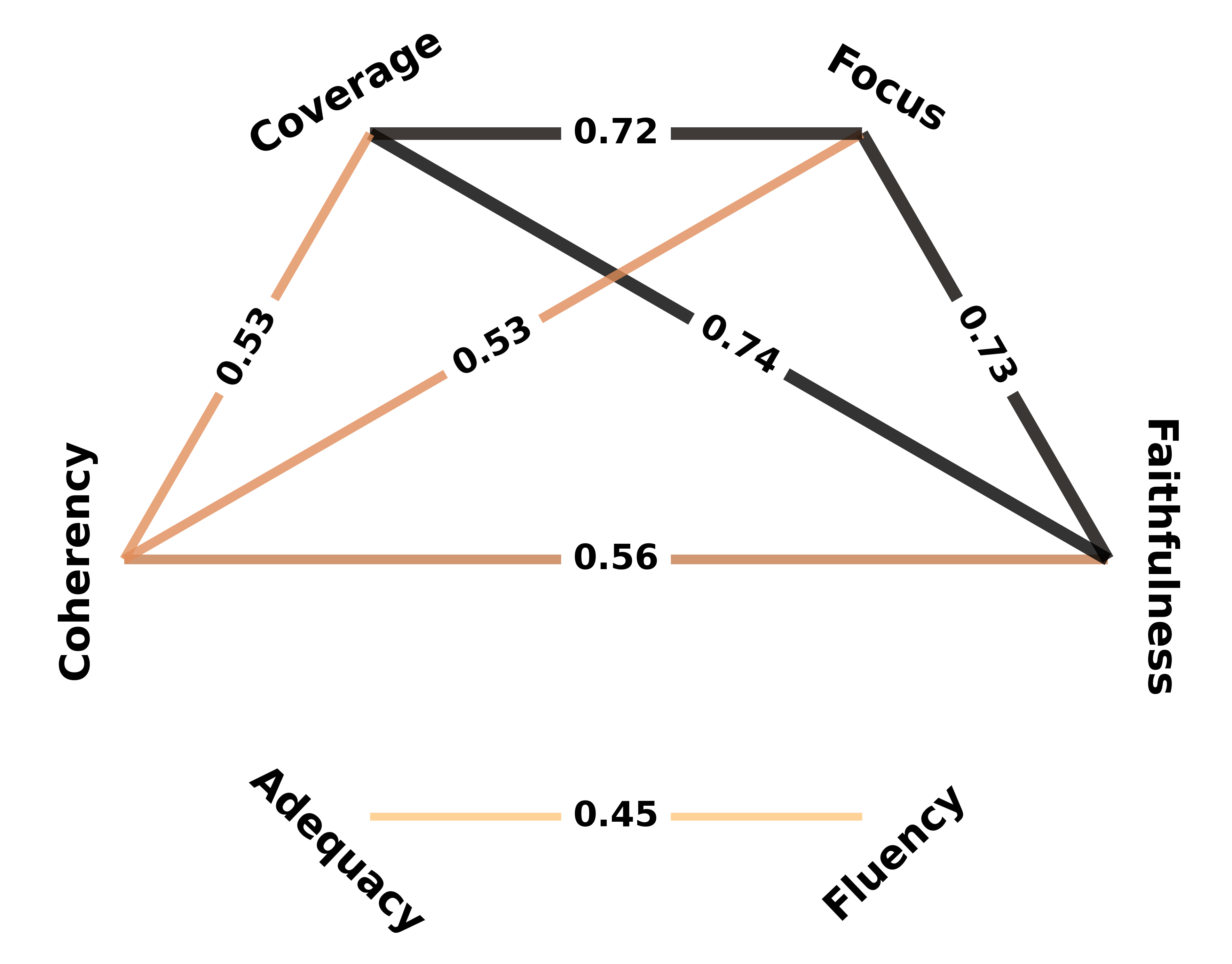}
  \end{center}
  \caption{Pearson correlation network of human evaluation aspects.}
  \label{fig:Aspect_Corr}
\end{figure}

From Figure~\ref{fig:human_scores}, we observe that the distributions of \textit{Faithfulness}, \textit{Focus}, and \textit{Coverage} are largely similar. Consistently, Figure~\ref{fig:Aspect_Corr} highlights their strong inter-metric correlations, whereas \textit{Coherency} exhibits notably weaker association. Meanwhile, \textit{Adequacy} and \textit{Fluency} show limited mutual dependence, reflecting their distinct evaluative roles.

\section{Experiments}

\subsection{Evaluation Metrics}
\label{sec:eval_metric}
To achieve a holistic and nuanced assessment, we employ a diverse collection of metrics that span surface-level text similarity, semantic understanding, learned human-aligned judgments, and cutting-edge LLM-based evaluation.

\noindent\textbf{Lexical and n-gram Overlap Metrics.} These metrics quantify direct textual correspondences between hypotheses and references. Our set includes BLEU (1-4) \citep{papineni-etal-2002-bleu}, sentBLEU \cite{chen2014systematic}, SacreBLEU \cite{post-2018-call}, GLEU \cite{mutton-etal-2007-gleu}, NIST \cite{doddington2002automatic}, LEPOR \cite{han-etal-2012-lepor}, ChrF \cite{popovic-2015-chrf}, ChrF++ \cite{popovic-2017-chrf}, METEOR \cite{banerjee-lavie-2005-meteor}, and a comprehensive suite of ROUGE variants \cite{lin-2004-rouge}. Specifically, we employ eight ROUGE variants—ROUGE-1, ROUGE-2, ROUGE-3, ROUGE-4, ROUGE-L, ROUGE-W, ROUGE-S4, and ROUGE-SU4. Each variant is evaluated under two stopword settings (with/without stopwords) and three scoring modes (precision, recall, F1-score), resulting in 48 distinct configurations.

\noindent\textbf{Embedding-based Semantic Metrics.} To capture semantic alignment beyond surface matches, we utilize embedding-based approaches, including BERTScore \cite{zhang2020bertscoreevaluatingtextgeneration}, BARTScore \cite{yuan2021bartscoreevaluatinggeneratedtext}, MoverScore \cite{zhao-etal-2019-moverscore}, Embedding Averaging \cite{landauer1997solution}, Greedy Matching \cite{rus-lintean-2012-comparison}, Vector Extrema \cite{forgues2014bootstrapping}, LaBSE \cite{feng-etal-2022-language}, and LASER \cite{artetxe2019massively}. BERTScore is computed using four models: IndicBERT \cite{kakwani-etal-2020-indicnlpsuite}, mBERT cased, mBERT uncased \cite{devlin-etal-2019-bert}, and XLM-RoBERTa \cite{conneau-etal-2020-unsupervised}, across all 12 layers and using precision, recall, and F-score, yielding 144 distinct configurations.

\noindent\textbf{Neural Learned Metrics.} Metrics specifically trained to emulate human judgment include BLEURT \cite{sellam-etal-2020-bleurt}, COMET \cite{rei-etal-2020-comet}, and xCOMET \cite{guerreiro-etal-2024-xcomet}, offering refined evaluation of nuanced language understanding.

\noindent\textbf{LLM-based Evaluation.} At the forefront, we harness the reasoning and discourse-aware capabilities of large language models for evaluation.
GPTScore \cite{fu-etal-2024-gptscore}, implemented with Flan-T5-XXL, evaluates outputs along six predefined dimensions, providing diagnostic insights that extend beyond overall scores. We additionally report MetricX-24 \cite{juraska-etal-2024-metricx} and GEMBA \cite{kocmi-federmann-2023-large}, with GEMBA instantiated using GPT-4.1. Complementing this, we utilize state-of-the-art LLMs—including GPT-4.1, LLaMA-3.3 \cite{grattafiori2024llama3herdmodels}, DeepSeek V3 \cite{deepseekai2025deepseekv3technicalreport}, and Gemini 2.5 \cite{comanici2025gemini25pushingfrontier}—through designed prompting protocols (full prompt details in Appendix~\ref{sec:app_llm_based}).

Full descriptions and implementation details for all metrics are available in Appendix~\ref{sec:app_eval_metric}.

\subsection{Outliers Detection}
We characterize outliers as observations that deviate markedly from the rest of the dataset \cite{barnett1994outliers}. Given Pearson correlation's sensitivity to such anomalies, we applied a robust detection method on the human scores for each evaluation aspect. Rather than relying on standardization, which is vulnerable to extreme values due to its dependence on the mean and standard deviation, we employed the median and Median Absolute Deviation (MAD) for greater resilience \cite{iglewicz1993volume, leys2013detecting}. For a set of human scores $s$, MAD was computed as:
\begingroup
\setlength{\abovedisplayskip}{4pt}
\setlength{\belowdisplayskip}{4pt}
\begin{equation*}
\text{MAD} = 1.483 \times \text{median}(|s - \text{median}(s)|)
\end{equation*}
\endgroup
A robust $z$-score was then derived using:
\begingroup
\setlength{\abovedisplayskip}{4pt}
\setlength{\belowdisplayskip}{4pt}
\begin{equation*}
z = \frac{s - \text{median}(s)}{\text{MAD}}
\end{equation*}
\endgroup
Observations with $|z| > 3.5$ were flagged as outliers \cite{mathur-etal-2020-tangled}. Table~\ref{tab:outliers_volume} reports the counts of detected outliers across languages and indicators.

\begin{table}[ht!]
\centering
\renewcommand{\arraystretch}{1}
\setlength{\tabcolsep}{6pt}
\resizebox{0.85\linewidth}{!}{%
\begin{tabular}{lcccccc}
\toprule
\multirow{2}{*}{\textbf{Language}} & \multicolumn{4}{c}{\textbf{Summarization}} & \multicolumn{2}{c}{\textbf{Translation}} \\
\cmidrule(lr){2-5} \cmidrule(lr){6-7}
 & \textbf{Fa.} & \textbf{Fo.} & \textbf{Cov.} & \textbf{Coh.} & \textbf{Ad.} & \textbf{Fl.} \\
\midrule
\rowcolor{gray!10} Hindi      & 0 & 0 & 0 & 0 & 0 & 0 \\
Bengali                         & 44 & 50 & 61 & 0 & 8 & 8 \\
\rowcolor{gray!10} Tamil       & 2 & 5 & 0 & 0 & 0 & 0 \\
Telugu                           & 0 & 0 & 0 & 0 & 0 & 0 \\
\rowcolor{gray!10} Marathi     & 0 & 0 & 0 & 0 & 0 & 3 \\
Gujarati                         & 37 & 45 & 0 & 0 & 20 & 22 \\
\bottomrule
\end{tabular}%
}
\caption{Outlier distribution across languages: Faithfulness (Fa), Focus (Fo), Coverage (Cov), Coherence (Coh), Adequacy (Ad), and Fluency (Fl).}
\label{tab:outliers_volume}
\end{table}

\subsection{Results and Analysis}

\subsubsection{Segment-level Analysis}

\begin{table*}[t!]
\centering
\scriptsize
\renewcommand{\arraystretch}{1.1}
\resizebox{1.0\textwidth}{!}{%
\begin{tabular}{l !{\vrule width 1pt} c c c c !{\vrule width 1pt} c c !{\vrule width 1pt} c}
\toprule
\multirow{2}{*}{\textbf{Metric}} & \multicolumn{4}{c!{\vrule width 1pt}}{\textbf{Summarization}} & \multicolumn{2}{c!{\vrule width 1pt}}{\textbf{Translation}} & \multirow{2}{*}{\textbf{Overall}} \\
\cmidrule(lr){2-5} \cmidrule(lr){6-7}
 & \textbf{Faithfulness} & \textbf{Focus} & \textbf{Coverage} & \textbf{Coherency} & \textbf{Adequacy} & \textbf{Fluency} &  \\
\midrule
BLEU-3 & 0.125 / 0.114 & 0.161 / 0.142 & 0.156 / 0.163 & 0.087 / 0.087 & 0.21 / 0.189 & 0.276 / 0.265 & 0.169 / 0.16 \\

sentBLEU      & 0.121 / 0.115 & 0.153 / 0.138 & 0.145 / 0.156 & 0.093 / 0.093 & 0.209 / 0.187 & 0.275 / 0.264 & 0.166 / 0.159 \\

SacreBLEU     & 0.074 / 0.051 & 0.09 / 0.064 & 0.068 / 0.058 & 0.013 / 0.013 & 0.138 / 0.132 & 0.192 / 0.184 & 0.096 / 0.084 \\

GLEU    & 0.079 / 0.078 & 0.113 / 0.099 & 0.103 / 0.114 & 0.085 / 0.085 & 0.206 / 0.184 & 0.27 / 0.259 & 0.143 / 0.137 \\

NIST    & 0.098 / 0.124 & 0.105 / 0.114 & 0.104 / 0.128 & 0.105 / 0.105 & 0.143 / 0.126 & 0.191 / 0.185 & 0.124 / 0.13 \\

METEOR    & 0.115 / 0.098 & 0.138 / 0.122 & 0.141 / 0.134 & 0.025 / 0.025 & 0.169 / 0.159 & 0.239 / 0.231 & 0.138 / 0.128 \\

LEPOR    & 0.187 / 0.175 & 0.167 / 0.166 & 0.19 / 0.195 & 0.075 / 0.075 & 0.1 / 0.08 & 0.095 / 0.096 & 0.136 / 0.131 \\

ROUGE-2-N-R    & 0.258 / \textbf{0.258} & 0.248 / 0.249 & \textbf{0.279} / \textbf{0.294} & 0.141 / 0.141 & 0.189 / 0.165 & 0.246 / 0.241 & 0.227 / 0.225 \\

ChrF    & 0.222 / 0.214 & 0.24 / 0.23 & 0.256 / 0.265 & 0.12 / 0.12 & 0.2 / 0.18 & 0.266 / 0.255 & 0.217 / 0.211 \\

ChrF++    & 0.205 / 0.191 & 0.23 / 0.214 & 0.239 / 0.245 & 0.116 / 0.116 & 0.201 / 0.181 & 0.269 / 0.257 & 0.21 / 0.201 \\

\hdashline
BERTScore    & 0.208 / 0.198 & 0.239 / 0.233 & 0.251 / 0.259 & 0.119 / 0.119 & 0.244 / 0.222 & 0.31 / 0.303 & 0.229 / 0.222 \\

BARTScore    & 0.106 / 0.13 & 0.121 / 0.134 & 0.119 / 0.132 & 0.05 / 0.05 & 0.058 / 0.068 & 0.108 / 0.11 & 0.094 / 0.104 \\

MoverScore    & 0.09 / 0.093 & 0.094 / 0.098 & 0.109 / 0.065 & 0.094 / 0.094 & 0.059 / 0.068 & 0.107 / 0.117 & 0.092 / 0.089 \\

Embedding Averaging    & 0.12 / 0.1 & 0.196 / 0.175 & 0.152 / 0.151 & 0.047 / 0.047 & 0.259 / 0.237 & 0.305 / 0.297 & 0.18 / 0.168 \\

Greedy Matching    & 0.121 / 0.101 & 0.199 / 0.177 & 0.152 / 0.151 & 0.048 / 0.048 & 0.259 / 0.236 & 0.307 / 0.298 & 0.181 / 0.169 \\

Vector Extrema    & 0.121 / 0.101 & 0.199 / 0.177 & 0.152 / 0.151 & 0.048 / 0.048 & 0.259 / 0.236 & 0.307 / 0.298 & 0.181 / 0.169 \\

LaBSE    & 0.132 / 0.112 & 0.185 / 0.169 & 0.172 / 0.172 & 0.081 / 0.081 & 0.286 / 0.264 & 0.346 / 0.34 & 0.2 / 0.19 \\

LASER    & 0.048 / 0.086 & 0.104 / 0.129 & 0.086 / 0.079 & 0.06 / 0.06 & 0.093 / 0.114 & 0.132 / 0.147 & 0.087 / 0.103 \\

\hdashline
BLEURT    & 0.175 / 0.188 & 0.201 / 0.208 & 0.223 / 0.226 & 0.163 / 0.163 & 0.291 / 0.289 & 0.362 / 0.361 & 0.236 / 0.239 \\

COMET    & 0.214 / 0.24 & 0.24 / 0.262 & 0.251 / 0.27 & 0.204 / 0.204 & 0.267 / 0.255 & 0.339 / 0.339 & 0.253 / 0.262 \\

xCOMET   &  -0.032 / -0.026 & -0.006 / -0.006 & -0.025 / -0.02  & -0.006 / -0.006 &  0.286 / 0.281 & 0.334 / 0.338 &  0.092 / 0.094  \\

\hdashline
GPTScore    & -0.053 / -0.044 & -0.074 / -0.048 & -0.062 / -0.047 & -0.048 / -0.048 & 0.008 / -0.011 & -0.003 / -0.009 & -0.039 / -0.035 \\

LLama-3.3-Instruct    & 0.228 / 0.219 & 0.252 / 0.242 & 0.24 / 0.244 & 0.159 / 0.159 & 0.263 / 0.232 & 0.335 / 0.334 & 0.246 / 0.238 \\

DeepSeek-V3    & 0.251 / 0.234 & \textbf{0.289} / 0.269 & 0.277 / 0.275 & 0.233 / 0.233 & \textbf{0.311} / 0.282 & \textbf{0.379} / \textbf{0.364} & \textbf{0.29} / \textbf{0.276} \\

GPT-4.1 \tiny(English prompt)   & \textbf{0.263} / 0.257 & 0.278 / 0.271 & 0.238 / 0.244 & \textbf{0.24} / \textbf{0.24} & 0.274 / 0.241 & 0.354 / 0.343 & 0.275 / 0.266 \\

GPT-4.1 \tiny(Native prompt)    & 0.235 / 0.242 & 0.241 / 0.254 & 0.245 / 0.254 & 0.168 / 0.168 & 0.289 / 0.262 & 0.359 / 0.347 & 0.256 / 0.255 \\

Gemini 2.5 Flash    & 0.248 / 0.244 & 0.277 / \textbf{0.274} & 0.257 / 0.264 & 0.202 / 0.202 & 0.295 / 0.265 & 0.363 / 0.347 & 0.274 / 0.266 \\

MetricX-24 \tiny(reference-free) & -/- & -/- & -/- & -/- & \textbf{0.311} / \textbf{0.298}  & 0.375 / 0.361 & 0.343 / 0.33\\

GEMBA \tiny(reference-free) & -/- & -/- & -/- & -/- & 0.295 / 0.263 & 0.366/ 0.354 & 0.331 / 0.309\\
\bottomrule
\end{tabular}
}
\caption{Segment-level Pearson correlations between automatic metrics and human judgments; values after `/` reflect results with outliers removed.}
\label{tab:segment-level}
\end{table*} 

Table~\ref{tab:segment-level} presents our segment-level analysis, highlighting Pearson correlations between automatic metrics and human judgments.

We identified the optimal configurations for three key metrics based on segment-level correlations with human assessments. Out of 48 ROUGE configurations explored, ROUGE-2 with recall scoring and stopwords removed emerged as the top performer. For BERTScore, IndicBERT at layer 8 with recall mode yielded the best results, while BLEU-3 was the most effective BLEU configuration. All subsequent references to these metrics adopt these high-performing settings.

Performance trends varied across metric categories. Lexical overlap metrics showed task-specific strengths: ROUGE excelled in TS, whereas BLEU-3 led in MT. Among embedding-based metrics, BERTScore was strongest for TS, while LaBSE dominated in MT. Neural learned metrics exhibited similar task preferences, with COMET outperforming in TS and BLEURT aligning better with MT human judgments.

LLM-based evaluators, however, displayed a consistent dominance. DeepSeek-V3 achieved the highest correlations for both TS and MT, establishing itself as the most reliable metric overall. Prompting also influenced performance: native-language prompts outperformed English in MT, whereas English prompts generally fared better in TS, except for \textit{Coverage}. Overall, LLM-based and neural learned metrics proved more dependable at the segment level than traditional metrics. A detailed comparison at the system level is provided in Appendix~\ref{sec:app_system_level_analysis}.

A complementary pattern emerges between summarization and translation evaluation. In TS, automatic metrics align closely with content-oriented dimensions like \textit{Coverage} and \textit{Focus}, but are less sensitive to \textit{Coherency}, highlighting limitations in capturing logical flow and overall readability. Conversely, MT evaluation prioritizes \textit{Fluency}, with \textit{Adequacy} reasonably captured, while finer semantic and discourse properties remain less reliably assessed. This suggests summarization metrics better track content preservation, whereas translation metrics better reflect how content is expressed, emphasizing a trade-off between content fidelity and linguistic quality.

Outliers significantly affect human–metric correlations, though their impact differs across indicators. In TS, removing outliers generally reduces correlations for \textit{Faithfulness} and \textit{Focus}, indicating previous inflation due to extreme samples. LASER, however, shows a substantial increase (+79.17\% for \textit{Faithfulness}, +24.04\% for \textit{Focus}\footnote{Percent changes quantify outlier effects as the relative difference between correlations with and without outliers, expressed as a percentage of the original correlation.}), suggesting prior underestimation. In contrast, \textit{Coverage} typically improves across most metrics following outlier removal, with a few exceptions—such as MoverScore and SacreBLEU. \textit{Coherency} remains unaffected, consistent with the absence of outliers reported in Table~\ref{tab:outliers_volume}.

For MT, excluding outliers generally decreases correlations for both \textit{Adequacy} and \textit{Fluency}, except for metrics like LASER, BARTScore, and MoverScore, which show improved alignment with human judgments.

Sensitivity to outliers varies widely across metrics. LASER (+18.39\%), BARTScore (+10.64\%), and SacreBLEU (-12.5\%) exhibit substantial shifts, whereas robust metrics such as ROUGE (-0.88\%) and GPT-4.1\small(Native prompt)\normalsize \, (-0.39\%) remain stable. Overall, embedding-based metrics are more vulnerable to outliers, while neural and LLM-based evaluators demonstrate consistent resilience.

Crucially, to our knowledge, this study is the first to systematically investigate the influence of outliers on human–metric correlations within Indian languages. Previous research in this area has not addressed this issue, underscoring both the novelty and the importance of our findings.

\subsubsection{Language Effects}

\begin{figure}[t!]
  \includegraphics[width=\columnwidth]{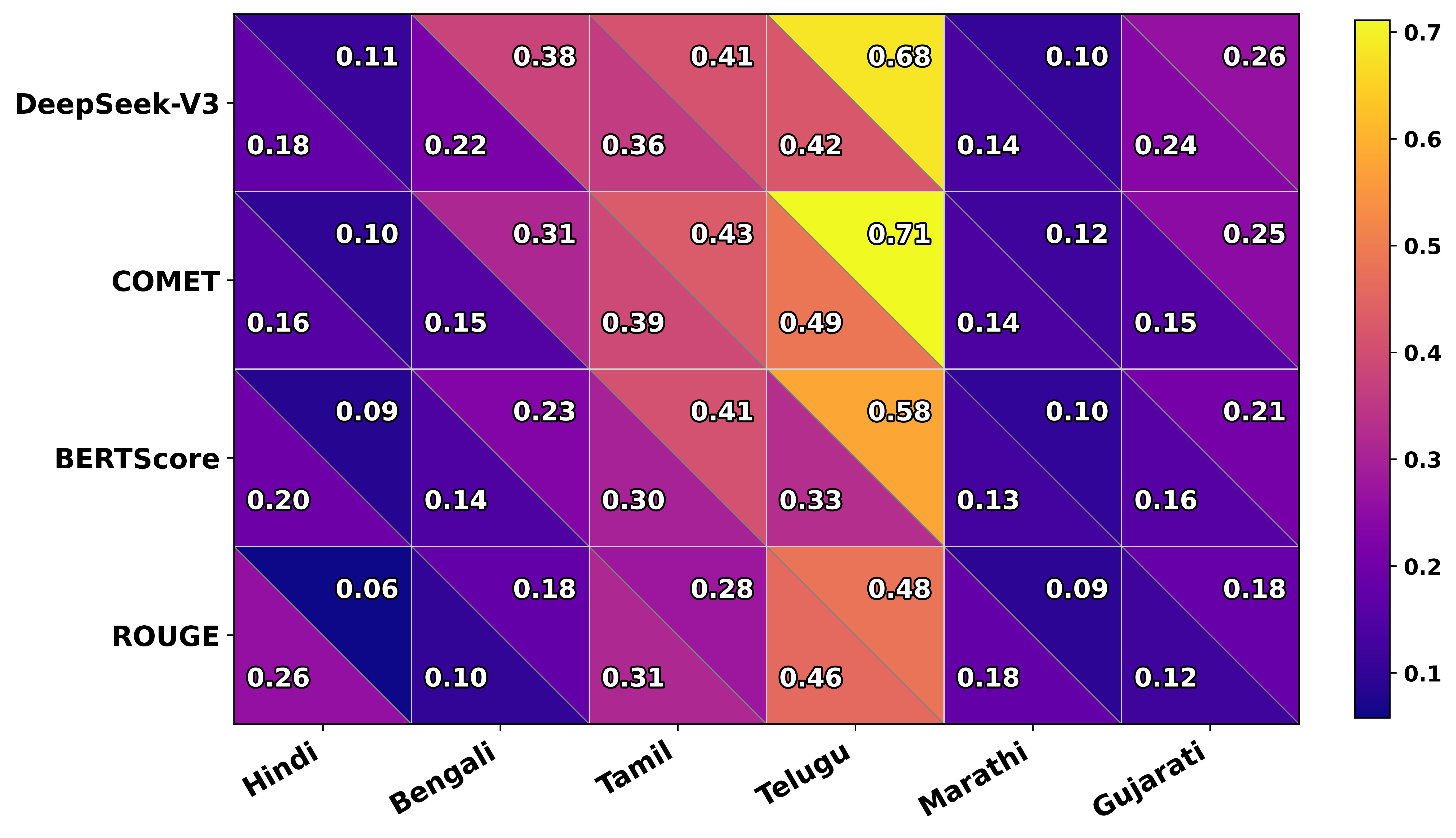}
  \caption{Language-specific Pearson correlations of top metrics across tasks (top-right: MT, bottom-left: TS).}
  \label{fig:LangWiseHeatmap}
\end{figure}

Figure~\ref{fig:LangWiseHeatmap} illustrates language-specific Pearson correlations for the top-performing metrics across both tasks, highlighting clear variations in performance across languages. Telugu consistently achieves the highest scores, whereas Marathi in TS and Hindi in MT rank lowest. Interestingly, LLM-based evaluations follow the same pattern. Notably, Hindi—a high-resource language—is evaluated less reliably than lower-resource languages like Tamil and Telugu, revealing an unexpected anomaly in LLM-based assessments that contrasts with previous evidence of superior performance in high-resource settings~\cite{li2025language, yari-koto-2025-unveiling}.

\subsubsection{Metric Inter-correlation}
\begin{figure*}[t!]
    \centering
    \begin{subfigure}{0.475\textwidth}
        \centering
        \includegraphics[width=\linewidth]{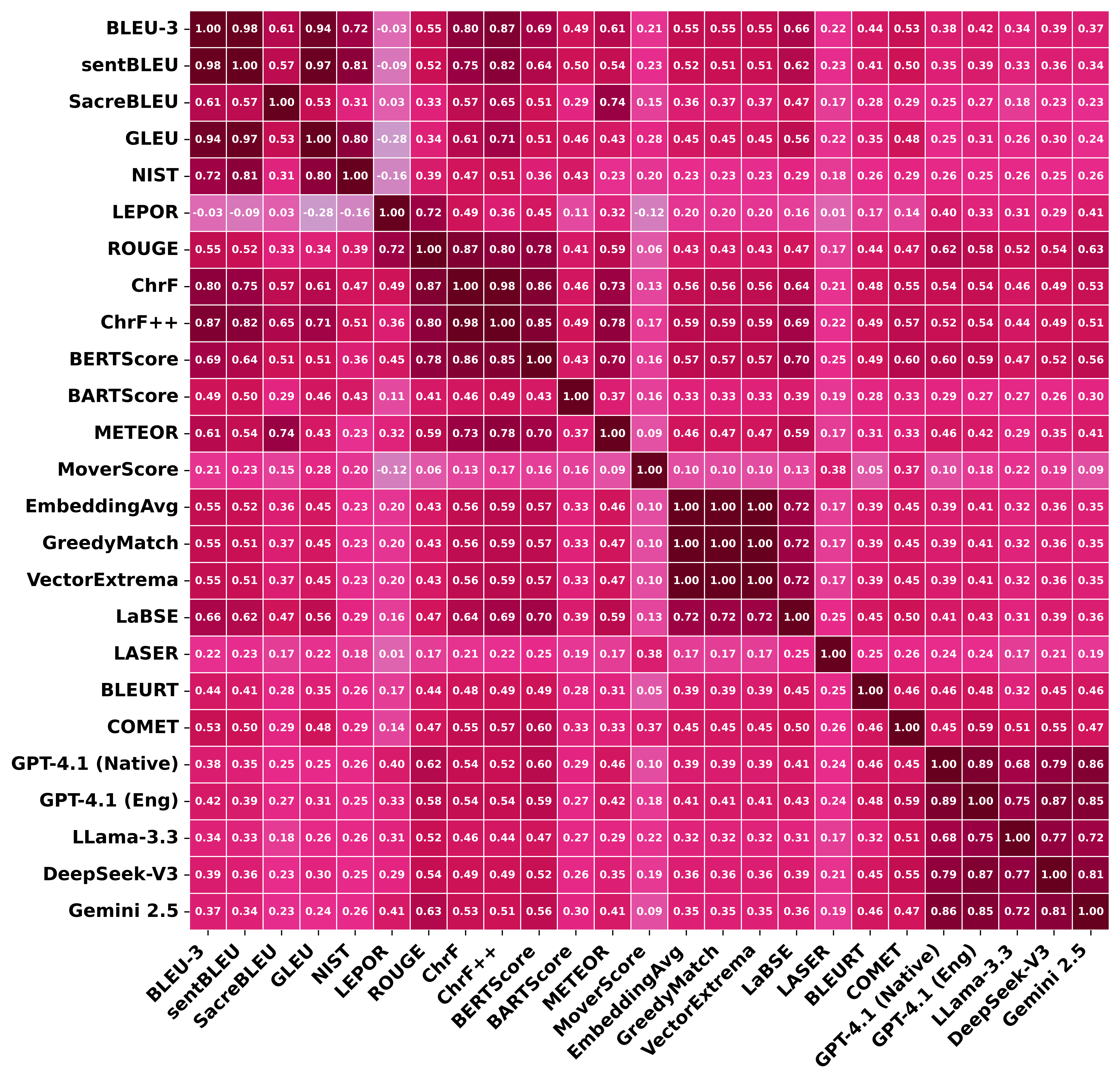}
        \caption{Text Summarization}
        \label{fig:summ_metric_corr}
    \end{subfigure}
    \hfill
    \begin{subfigure}{0.475\textwidth}
        \centering
        \includegraphics[width=\linewidth]{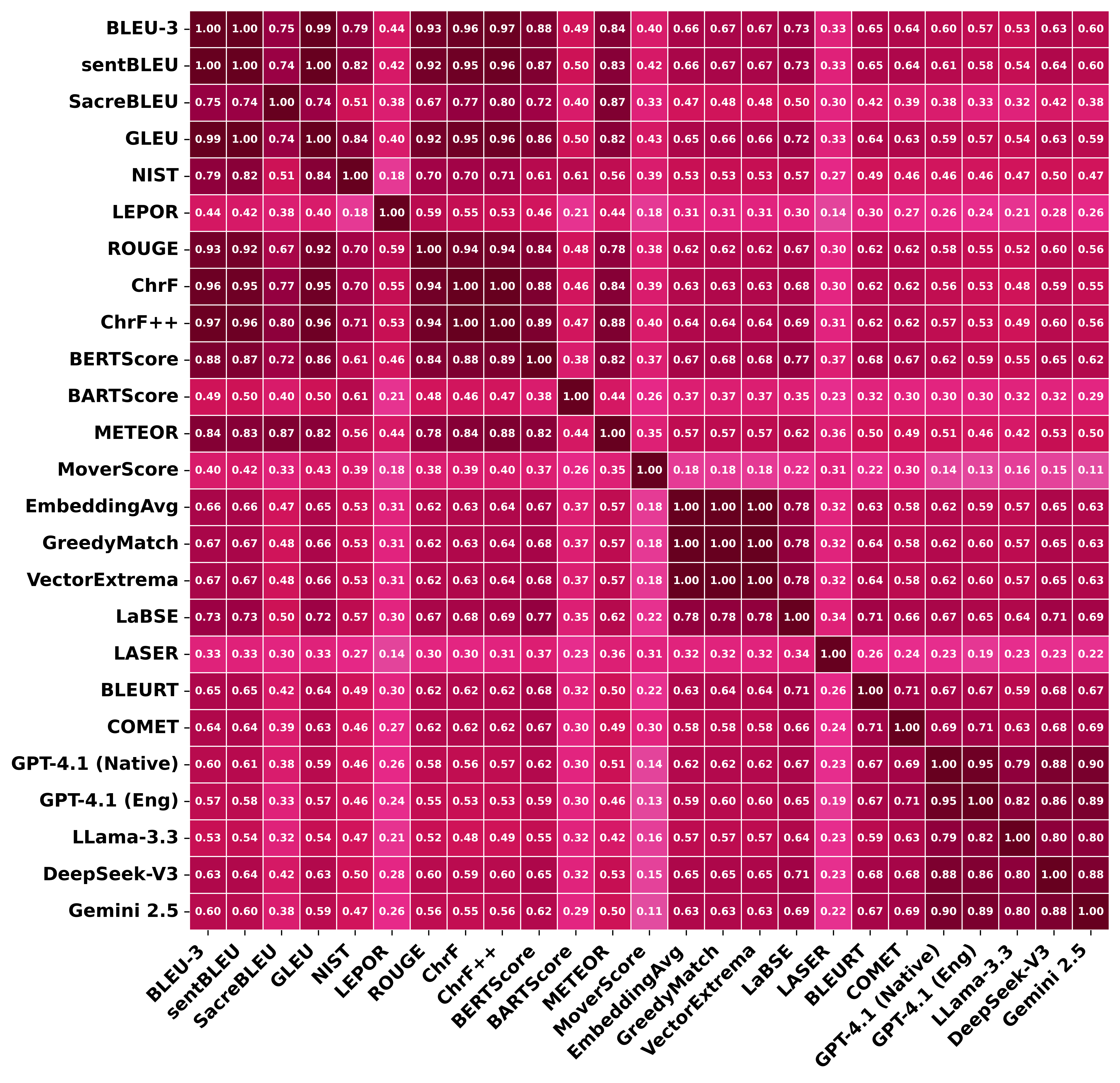}
        \caption{Machine Translation}
        \label{fig:trans_metric_corr}
    \end{subfigure}
    \caption{Pearson correlation matrices of automatic metrics.}
    \label{fig:metric corr}
\end{figure*}

Figure~\ref{fig:metric corr} presents the Pearson correlation matrices of automatic metrics, uncovering well-defined clusters that reflect our proposed taxonomy. Lexical overlap metrics formed a compact, highly correlated group, reaffirming their common reliance on surface-level n-gram matching. Embedding-based metrics likewise emerged as a cohesive cluster, while vector-based metrics displayed perfect internal agreement.

LLM-based evaluators distinguished themselves with strikingly high mutual correlations ($r \approx 0.7{-}0.9$) across both tasks, underscoring their consistent judgment patterns despite architectural diversity. Yet, their only moderate alignment with traditional metrics underscores a methodological divide between evaluation grounded in large language models and established paradigms.

Notably, LEPOR, MoverScore, and LASER persistently showed the weakest average correlations with other metrics, indicating either that they capture unique quality dimensions or diverge from the prevailing consensus of evaluation methods.

\subsection{Robustness \& Sensitivity}

\subsubsection{Paraphrasing Sensitivity}
\begin{figure}[ht!]
  \begin{center}
     \includegraphics[width=0.7\columnwidth]{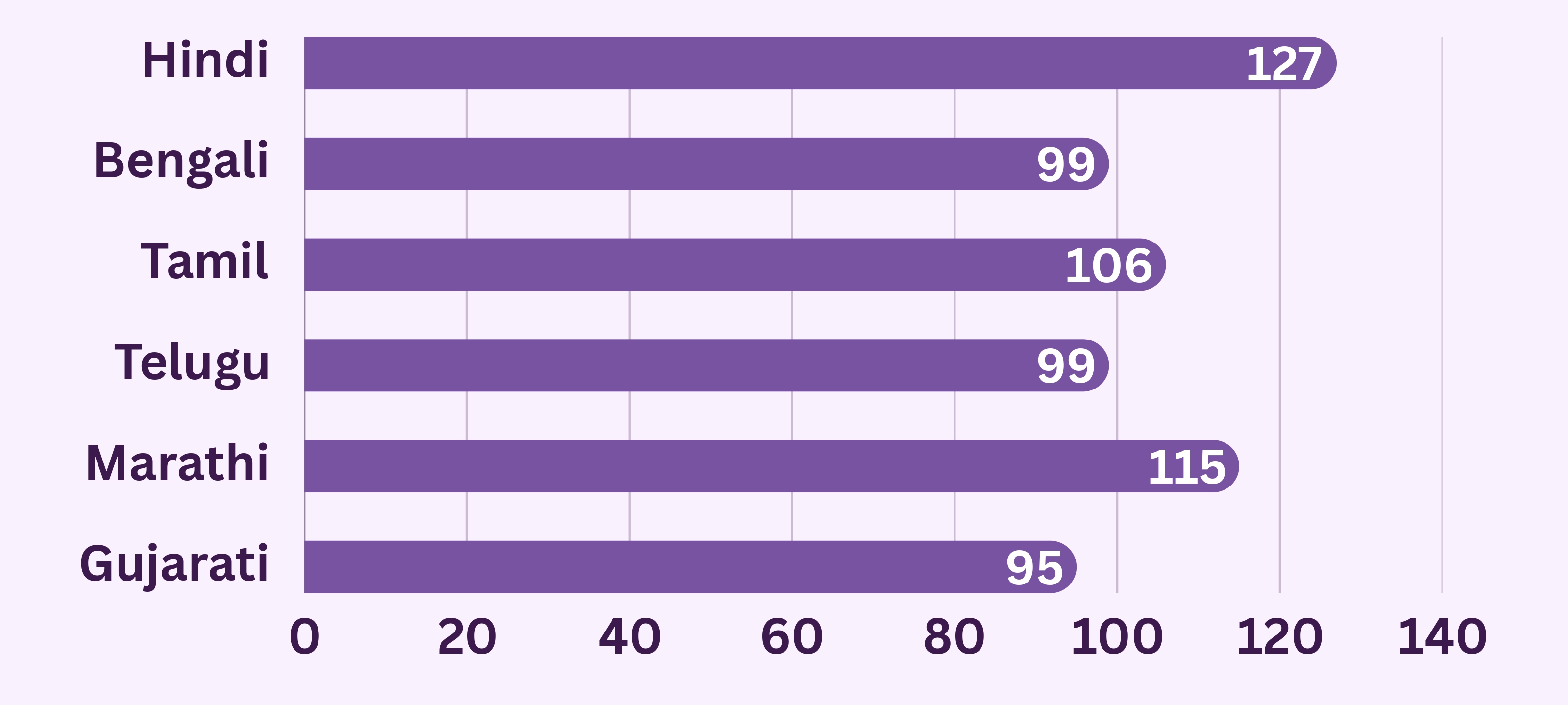}
  \end{center}
  \caption{Distribution of high-quality paraphrased summaries across languages.}
  \label{fig:Paraphrased_Count}
\end{figure}

To evaluate the robustness of automatic metrics beyond surface word overlap, we conducted a controlled paraphrasing experiment. GPT-4.1 rephrased human-written summaries (prompt in Appendix~\ref{sec:app_paraphrasing_prompt}), followed by a two-step validation. In each language, two native speakers assessed the outputs based on four criteria: meaning preservation, grammaticality, clarity, and detail consistency. A primary annotator first reviewed and minimally corrected borderline cases, after which a second annotator verified quality, retaining only paraphrases with unanimous approval. Figure~\ref{fig:Paraphrased_Count} shows the number of validated paraphrases out of 150 per language.

\subsubsection{Noise Perturbations}
We evaluated metric resilience under controlled perturbations affecting semantic fidelity, surface fluency, and summarization-specific traits. By introducing noise to reference outputs, we simulated real-world imperfections to assess each metric’s reliability and sensitivity across linguistic dimensions.

\noindent\textbf{Semantic Perturbations} focused on meaning alterations. Using IndicNER—a multilingual named entity recognition model for Indian languages \cite{mhaske-etal-2023-naamapadam}—we constructed lexicons of persons, locations, and organizations. For each text, a randomly selected entity was substituted, deleted, corrupted at the character level, or augmented with an additional entity of the same type. Further semantic modifications included numeric adjustments, verb negation using language-specific markers, and synonym or antonym replacements. These manipulations primarily affect MT \textit{Adequacy} and TS \textit{Faithfulness}.

\noindent\textbf{Lexical and Structural Perturbations} targeted MT \textit{Fluency} and TS \textit{Coherency}. In translations, words were shuffled within a four-token window, whereas summaries experienced sentence-level shuffling through pairwise swaps. Additional noise included random stopword insertions or deletions, as well as character-level edits\footnote{replacement, insertion, deletion, or transposition}.

\noindent\textbf{Summarization-specific Perturbations} probed higher-level content properties. To evaluate \textit{Focus}, we added a contextually plausible but off-topic sentence from another summary. To challenge \textit{Coverage}, we removed the sentence with the lowest ROUGE overlap, simulating the loss of the most informative content.

\begin{figure}[t]
  \includegraphics[width=\linewidth]{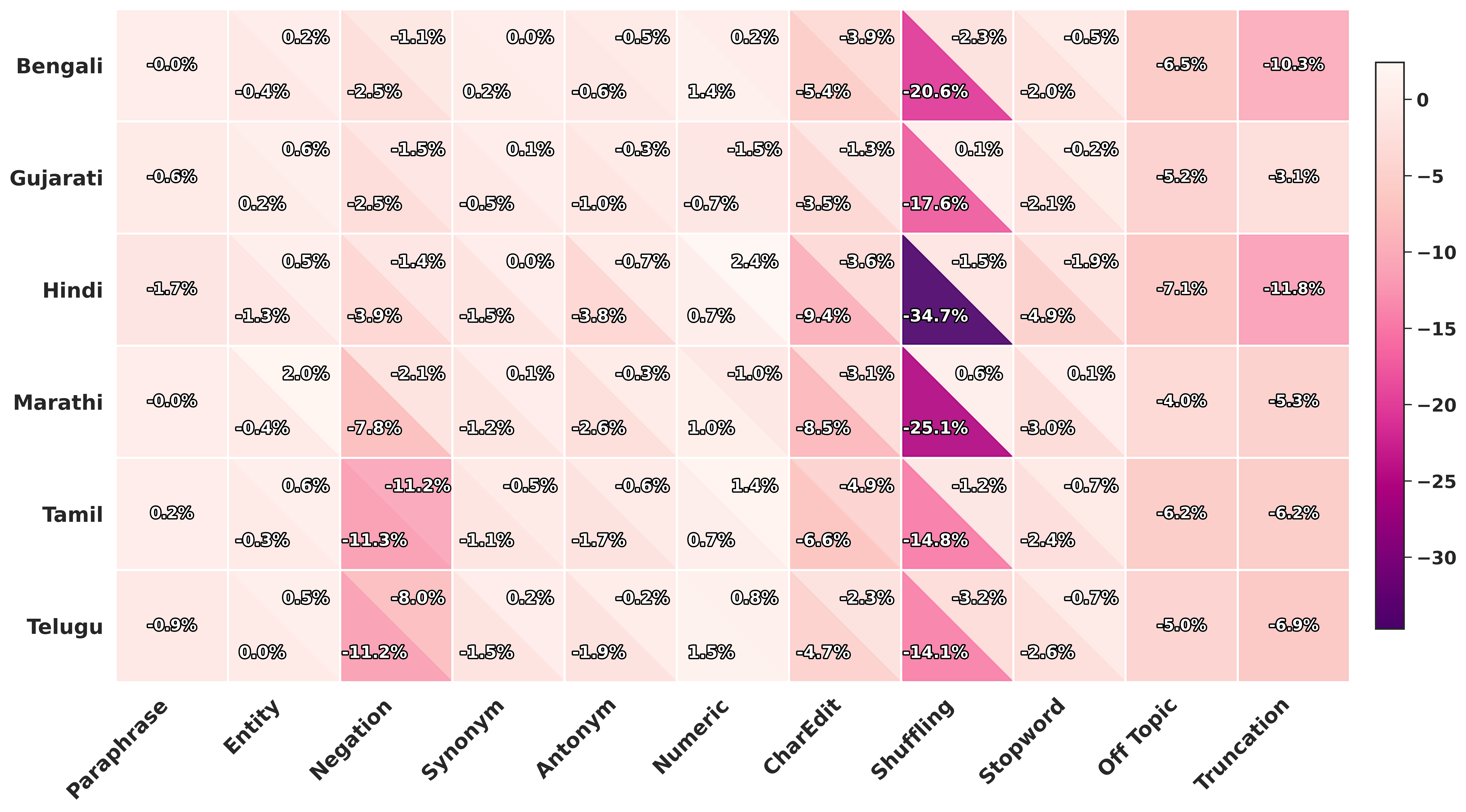}
  \vspace{-0.2cm}
  \caption{Relative COMET score drops (\%) across languages under diverse perturbations (top-right: TS; bottom-left: MT).}
  \label{fig:COMET_Drop_Lang}
\end{figure}

To describe the results, we focus on COMET as one of the most reliable open-source evaluation metrics, with complementary analyses for ROUGE and BERTScore provided in Appendix~\ref{sec:app_additional_metric_senitivity}. Figure~\ref{fig:COMET_Drop_Lang} shows COMET’s relative score drops across languages under different perturbations. In MT, shuffling is the most disruptive, while in TS, truncating key sentences leads to the greatest decline. Paraphrasing and synonym replacements have minimal impact, indicating COMET’s robustness to meaning-preserving variations. Negation effects vary by language: Tamil and Telugu are highly sensitive, while others remain stable. Overall, Hindi is most vulnerable and Gujarati most resilient.

\begin{figure}[t]
  \includegraphics[width=\linewidth,keepaspectratio]{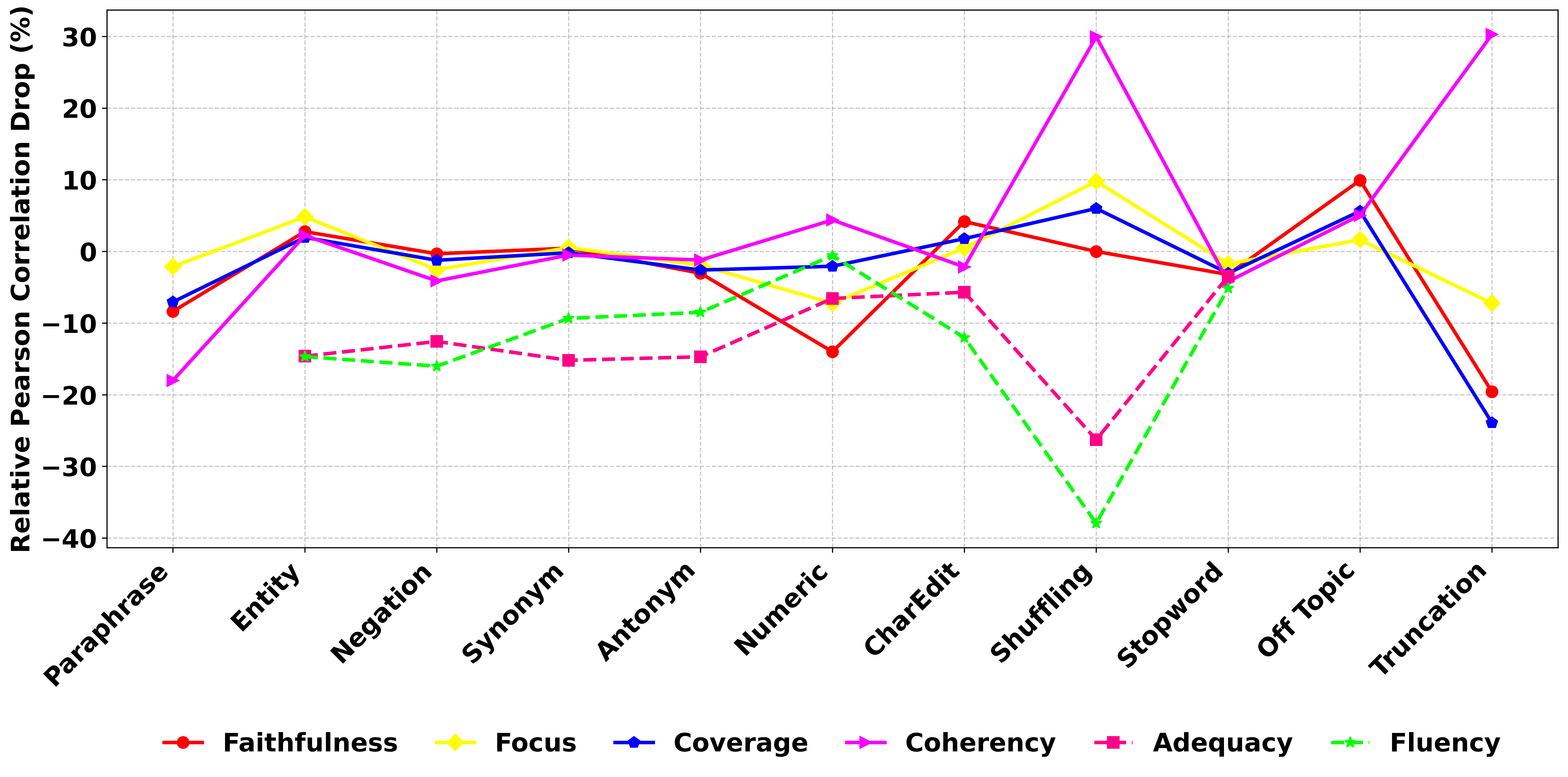}
  \vspace{-0.2cm}
  \caption{Relative Pearson correlation drop (\%) between human judgments and COMET across evaluation aspects under perturbations.}
  \label{fig:COMET_Corr_Aspect}
\end{figure}

Figure~\ref{fig:COMET_Corr_Aspect} presents the relative declines in human–COMET correlations across evaluation aspects. In MT, shuffling causes the sharpest drops in \textit{Adequacy} and \textit{Fluency}; in TS, truncation reduces correlations for \textit{Faithfulness}, \textit{Focus}, and \textit{Coverage}, but unexpectedly increases \textit{Coherency}. Overall, \textit{Focus} remains the most stable aspect, whereas \textit{Fluency} is least robust. These findings highlight COMET’s strengths in handling lexical variation but also its limitations under structural and meaning-altering perturbations, reinforcing the importance of targeted robustness evaluations across languages and quality dimensions.

\section{Conclusion}
This study introduces \texttt{\textbf{ITEM}}, a benchmark for evaluating automatic metrics in MT and TS across six Indian languages and quality dimensions, including alignment with human judgments, outlier effects, language consistency, inter-metric correlations, and robustness under perturbations. Results show that LLM-based evaluators correlate most strongly with human ratings, with metrics emphasizing fidelity in TS and fluency in MT. Robustness analyses reveal vulnerabilities to structural and semantic noise, highlighting the need for more resilient evaluation methods. Overall, \texttt{\textbf{ITEM}} provides a foundation for developing future metrics that are reliable and interpretable for Indian languages.

\section*{Limitations}
Despite offering a comprehensive framework for evaluating MT and TS metrics in Indian languages, this study is not without limitations. Our analysis is confined to six widely spoken languages, leaving out many low-resource Indian languages whose rich typological variation could divulge further vulnerabilities in existing metrics. Moreover, while LLM-based evaluators showed the strongest alignment with human judgments, their reliance on resource-heavy, opaque, and often proprietary systems raises concerns about accessibility, reproducibility, and interpretability—highlighting the need for developing transparent and efficient alternatives. Finally, although our robustness experiments covered a diverse set of controlled perturbations, they cannot fully capture the complexity of real-world textual noise. 

Future work should explore hybrid strategies that combine human-in-the-loop feedback, linguistic knowledge, and LLM reasoning to build more reliable and inclusive multilingual evaluation frameworks.

\bibliography{custom}

\appendix

\begin{table*}[t]
\centering
\setlength{\tabcolsep}{4pt}
\renewcommand{\arraystretch}{1.2}
\resizebox{0.9\textwidth}{!}{%
\begin{tabularx}{\textwidth}{l c c c c c}
\toprule
\textbf{Reference} &
\textbf{Lang. Coverage\textsuperscript{\P}} &
\textbf{Tasks} &
\textbf{\# Metrics} &
\textbf{Outliers\textsuperscript{\textdagger}} &
\textbf{Robustness\textsuperscript{\S}} \\
\midrule
\citet{bhattacharjee-etal-2023-crosssum} & hi, bn, ta, mr, gu & TS & 2 & \xmark & \xmark \\
\citet{sai-b-etal-2023-indicmt} & hi, ta, mr, gu & MT & 15 & \xmark & \cmark \\
\citet{gala2023indictrans2highqualityaccessiblemachine} & hi, bn, ta, te, mr, gu & MT & 3 & \xmark & \xmark \\
\citet{singh-etal-2024-good}  & -- & MT & 15 & \xmark & \xmark \\
\citet{hada-etal-2024-metal} & hi, bn & TS & 3 & \xmark & \xmark \\
\citet{han-etal-2024-rethinking} & hi, bn, ta & TS & 6 & \xmark & \xmark \\
\citet{deutsch-etal-2025-wmt24} & hi, bn, ta, te, mr, gu & MT & 6 & \xmark & \xmark \\
\citet{kocmi-etal-2025-findings} & hi, bn, mr & MT & 5 & \cmark & \cmark \\

\midrule
ITEM (Ours) & hi, bn, ta, te, mr, gu & MT+TS & 29 & \cmark & \cmark \\
\bottomrule
\end{tabularx}%
}
\caption{Comparison of Indic evaluation resources.
\textsuperscript{\P}\ \textit{Lang. Coverage:} languages jointly covered.
\textsuperscript{\textdagger}\ \textit{Outliers:} whether explicit outlier handling is reported.
\textsuperscript{\S}\ \textit{Robustness:} whether explicit metric robustness analysis via perturbations is reported.}
\label{tab:datasets_comparison}
\vspace{-0.0cm}
\end{table*}

\section{Indic Evaluation Benchmarks}
We situate ITEM within existing Indic evaluation resources for machine translation and summarization. Table~\ref{tab:datasets_comparison} compares prior benchmarks by shared language coverage, supported tasks, metric breadth, and whether studies explicitly report outlier handling and robustness analyses.

\section{Evaluation Metrics}
\label{sec:app_eval_metric}

\subsection{Lexical and n-gram Overlap Metrics}
\paragraph{BLEU:} The Bilingual Evaluation Understudy (BLEU) quantifies the similarity between a generated hypothesis and a reference translation. It captures adequacy through word-level precision and fluency via n-gram precision across n = 1 to 4. BLEU computation integrates three components: (1) n-gram precision comparing the output with reference translations, (2) a brevity penalty (BP) to avoid inflated scores from overly short sentences, and (3) clipping to regulate repeated word occurrences.

\paragraph{SentBLEU:} A sentence-level adaptation of the BLEU metric, evaluates each sentence independently rather than averaging n-gram precisions across an entire corpus, offering a more granular assessment of machine-generated quality.

\paragraph{SacreBLEU:} SacreBLEU provides a standardized, reproducible variant of the BLEU metric, ensuring consistent evaluation across studies. By enforcing uniform tokenization, normalization, and punctuation handling, it mitigates discrepancies arising from differing preprocessing or parameter choices in traditional BLEU implementations. 

\paragraph{GLEU:} Google BLEU (GLEU) adapts BLEU for sentence-level evaluation, making it ideal for assessing individual machine-generated outputs. It measures the overlap of n-grams between a candidate and a reference sentence, while applying a length penalty to discourage overgeneration, providing a more precise reflection of sentence-level quality.

\paragraph{NIST:} The metric, developed by the National Institute of Standards and Technology, evaluates similarity by assigning weights to matching n-grams according to their informational content. Rare n-gram sequences receive higher weights, reflecting their greater contribution to conveying meaningful content compared to more common sequences.

\paragraph{LEPOR:} LEPOR (Length Penalty, Precision, n-gram Position Difference Penalty, and Recall) is a robust, language-independent metric. It integrates multiple factors—an enhanced length penalty, precision, n-gram position difference penalty, and recall—without relying on external linguistic resources.

\paragraph{ROUGE Metrics:} The ROUGE (Recall-Oriented Understudy for Gisting Evaluation) suite quantifies the overlap between a machine-generated text and a reference, focusing on shared textual units such as n-grams or word sequences. Key variants include ROUGE-N (n-gram overlap), ROUGE-L (longest common subsequence), ROUGE-W (weighted LCS), ROUGE-S4 (skip-bigrams within a fixed window), and ROUGE-SU4 (skip-bigrams plus unigrams). Each metric offers precision, recall, and F-score measures, enabling a comprehensive assessment of summary quality.

\paragraph{ChrF / ChrF++:} The Character n-gram F-score (ChrF) evaluates the similarity between generated and reference texts at the character level, rather than using word n-grams. This approach reduces sensitivity to tokenization, making it robust across languages and text processing variations.

\paragraph{METEOR:} METEOR (Metric for Evaluation of Translation with Explicit ORdering) evaluates machine-generated text by combining a weighted F-score with a penalty function that balances precision and recall. The method begins by aligning the hypothesis with the reference to identify the longest matching sequences of words, incorporating synonyms from thesauri or large corpora to capture semantic equivalence and morphological variations. Precision and recall are computed based on matched unigrams, while the penalty function discourages short, fragmented matches and rewards longer, contiguous sequences. This design ensures a nuanced assessment of quality, emphasizing both accuracy and fluency.

\subsection{Embedding-based Semantic Metrics}
\paragraph{BERTScore:} Evaluates the similarity between a hypothesis and reference by leveraging contextual embeddings from a pre-trained BERT model, which is trained for masked language modeling and next sentence prediction. It computes the cosine similarity between the token embeddings of the machine-generated output and the reference, using greedy matching to maximize alignment. The metric reports precision, recall, and F1-score, providing a nuanced assessment of semantic overlap beyond surface-level matching.

\paragraph{BARTScore:} A text generation evaluation metric that leverages BART's encoder–decoder architecture to bridge the gap between training objectives and sentence generation. Trained with a denoising objective, BART is particularly effective for assessing conditional generation tasks. BARTScore computes the weighted log-probability of the reference sentence given the hypothesis, using the pre-trained parameters of BART, providing a precise measure of generation quality.

\paragraph{MoverScore:} Quantifies the semantic similarity between a summary and its reference by applying the Word Mover’s Distance on n-gram embeddings derived from BERT representations.

\paragraph{Embedding Averaging:} Represents a sentence by averaging the embeddings of all its words and measures sentence similarity using cosine similarity.

\paragraph{Greedy Matching:} Evaluates semantic alignment by pairing each word in the generated sentence with its closest semantic counterpart in the reference, producing an overall similarity score.

\paragraph{Vector Extrema:} Constructs sentence embeddings by taking the dimension-wise maxima or minima of word embeddings, emphasizing dominant semantic features. The cosine similarity between these vectors quantifies the semantic similarity between the generated and reference text.

\paragraph{LaBSE \& LASER:} Language-agnostic BERT Sentence Embeddings (LaBSE) and Language-Agnostic SEntence Representations (LASER) are pre-trained multilingual models designed to produce dense, language-independent sentence embeddings. These embeddings allow direct semantic comparison of sentences across diverse languages. LaBSE leverages a BERT-based architecture tailored for cross-lingual tasks, whereas LASER uses a BiLSTM encoder trained on parallel corpora to achieve its multilingual representations.

\subsection{Neural Learned Metrics}
\paragraph{BLEURT:} BLEURT (Bilingual Evaluation Understudy with Representations from Transformers) is a BERT-based, pre-trained model that leverages multi-task loss on extensive synthetic datasets comprising numerous reference sentences. Designed as a sentence-level evaluation metric, it produces prediction scores that quantify the similarity between a generated hypothesis and reference texts, effectively approximating human judgment. Its core objective is to train regression models capable of predicting human evaluation scores accurately.

\paragraph{COMET:} COMET (Cross-lingual Optimized Metric for Evaluation of Translation) is an evaluation framework that integrates ranking and regression techniques. It comprises an estimator and a reference-based ranking model grounded in human judgments. Employing a dual cross-lingual modeling approach for encoder inputs, COMET simultaneously captures information from both source and target sentences, enhancing performance. To emulate human evaluation, it adopts diverse training objectives: the estimator directly predicts the quality score of machine-generated text, while the ranking model optimizes by minimizing the distance between high-quality outputs and references, and maximizing the distance for lower-quality outputs. Both components rely on cross-lingual encoders, ensuring robust assessment across languages.

\subsection{LLM-based Evaluation}
\label{sec:app_llm_based}
\paragraph{GPTScore:} An evaluation framework leveraging generative pre-trained models, such as Flan-T5, to assess text quality. It operates on the premise that high-quality generated text is assigned higher probabilities by the model when conditioned on a given instruction and context. Unlike traditional form-filling approaches, GPTScore frames evaluation as a conditional generation task, capturing the nuanced likelihood of coherent and contextually appropriate outputs.

\paragraph{LLM-as-judges:} We harness the evaluative power of state-of-the-art LLMs—including GPT-4.1, LLaMA-3.3, DeepSeek V3, and Gemini 2.5—by designing prompts that enable them to function as reasoning-driven, discourse-aware evaluators. These models provide holistic assessments comparable to expert human reviewers. To maintain consistency and reproducibility, we utilize standardized prompt templates for both summarization and translation tasks.

\begin{promptbox}

\textsf{\textbf{System role:}} You are an expert \textit{\{TASK\}} evaluator. \\

\textsf{\textbf{Task:}} Rate the MACHINE \textit{\{SUMMARY/TRANSLATION\}} strictly on \textit{\{ASPECT\}}, defined as: \textit{\{ASPECT\_DEFINITION\}}. \\[1mm]

\textsf{\textbf{Scoring Rubric:}}\\
• Use a continuous scale from 0.0 (completely inadequate) to 1.0 (perfect).\\
• Any decimal allowed (e.g., 0.23, 0.46, 0.71)\\[1mm]

\textsf{\textbf{Instructions:}}\\
• Use {\scriptsize (ARTICLE / HUMAN SUMMARY for TS, SOURCE / HUMAN TRANSLATION for MT)} as a reference\\
• Evaluate only on \textit{\{ASPECT\}}—ignore other criteria\\
• Output exactly one numeric value between 0.0 (worst) and 1.0 (best).\\
• No extra text, rationale, or comments\\[1mm]

\textsf{\textbf{Input:}}\\[0.5mm]
\hspace*{2mm}ARTICLE / SOURCE\\
\hspace*{2mm}HUMAN SUMMARY / TRANSLATION\\
\hspace*{2mm}MACHINE SUMMARY / TRANSLATION\\[0.5mm]

\textsf{\textbf{Output:}}

\end{promptbox}

\section{Paraphrasing Prompt}
\label{sec:app_paraphrasing_prompt}

To guarantee transparency and reproducibility, we present the exact prompt employed with GPT-4.1 to produce paraphrased summaries.

\begin{promptbox}

\textsf{\textbf{System Role:}} You are an expert paraphraser specialized in rewriting text without changing its meaning. \\[0.5mm]

\textsf{\textbf{Task:}} Paraphrase the given HUMAN SUMMARY so that the core meaning remains unchanged, but the wording and structure are substantially different. \\[0.5mm]

\textsf{\textbf{Instructions:}}\\
• Preserve all factual and semantic content exactly.\\
• Use different vocabulary and sentence structures.\\
• Keep the paraphrase roughly the same length as the original.\\
• Maintain the same tone and style as the original.\\
• Do not add, remove, or infer any information.\\
• Do not explain your reasoning or include any comments.\\
• Output exactly one paraphrased summary, and nothing else. \\[0.5mm]

\textsf{\textbf{Input:}}\\
\hspace*{2mm}HUMAN SUMMARY: \{HUMAN\_SUMMARY\} \\[0.5mm]

\textsf{\textbf{Output:}}

\end{promptbox}

\section{Additional Analysis}
\label{sec:app_additional_analysis}

\subsection{System-level Analysis}
\label{sec:app_system_level_analysis}
Table~\ref{tab:system-level} reports Kendall-tau correlations at the system level, reflecting how well automatic metrics align with human judgments in ranking systems. Results reveal remarkable consistency, with most metrics achieving perfect correlations, underscoring their reliability in system ranking. Interestingly, the scores are uniform across all the dimensions assessed, reflecting a consistent concordance between metrics and human evaluations.

Only a handful of metrics—such as SacreBLEU, LEPOR, ROUGE, ChrF, NIST, and BARTScore—show minor deviations across the evaluated dimensions, yet the overall picture confirms the strong effectiveness of automatic metrics in capturing system-level performance.

\begin{table}[ht!]
\centering
\scriptsize
\renewcommand{\arraystretch}{1.2}
\setlength{\tabcolsep}{4pt}
\resizebox{0.48\textwidth}{!}{%
\begin{tabular}{l c c c}
\toprule
\textbf{Metric(s)} & \textbf{TS} & \textbf{MT} & \textbf{Overall} \\
\midrule
\multicolumn{4}{l}{\textit{Lexical Overlap}} \\
BLEU, sentBLEU, GLEU, METEOR, ChrF++ & \cellcolor{teal4}1.00 & \cellcolor{teal4}1.00 & \cellcolor{teal4}1.00 \\
SacreBLEU, LEPOR, ROUGE, ChrF & \cellcolor{teal1}0.33 & \cellcolor{teal4}1.00 & \cellcolor{teal2}0.55 \\
NIST & \cellcolor{teal4}1.00 & \cellcolor{teal1}0.33 & \cellcolor{teal3}0.78 \\
\hdashline
\multicolumn{4}{l}{\textit{Embedding-based}} \\
\begin{tabular}[t]{@{}l@{}}
BERTScore, MoverScore, Emb.~Avg,\\
Greedy, Vec.~Ext., LaBSE, LASER
\end{tabular} & \cellcolor{teal4}1.00 & \cellcolor{teal4}1.00 & \cellcolor{teal4}1.00 \\
BARTScore & \cellcolor{teal1}0.33 & \cellcolor{teal4}1.00 & \cellcolor{teal2}0.55 \\
\hdashline
\multicolumn{4}{l}{\textit{Neural Learned}} \\
xCOMET & \cellcolor{teal0}-0.33 & \cellcolor{teal4}1.00 & \cellcolor{teal1}0.33 \\
\begin{tabular}[t]{@{}l@{}}
BLEURT, COMET \end{tabular} & \cellcolor{teal4}1.00 & \cellcolor{teal4}1.00 & \cellcolor{teal4}1.00 \\
\hdashline
\multicolumn{4}{l}{\textit{LLM-based Evaluators}} \\
GPTScore & \cellcolor{teal0}-0.33 & \cellcolor{teal0}-0.33 & \cellcolor{teal0}-0.33 \\
\begin{tabular}[t]{@{}l@{}}
Llama-3.3, DeepSeek-V3, GPT-4.1(Eng/Native)\\
Gemini 2.5 Flash, GEMBA, MetricX-24
\end{tabular} & \cellcolor{teal4}1.00 & \cellcolor{teal4}1.00 & \cellcolor{teal4}1.00 \\
\bottomrule
\end{tabular}
}
\vspace{-0.1cm}
\caption{System-level Kendall correlations with human judgments.}
\vspace{-0.3cm}
\label{tab:system-level}
\end{table}

\subsection{Compression Effect}
\begin{figure}[ht!]
  \includegraphics[width=\columnwidth]{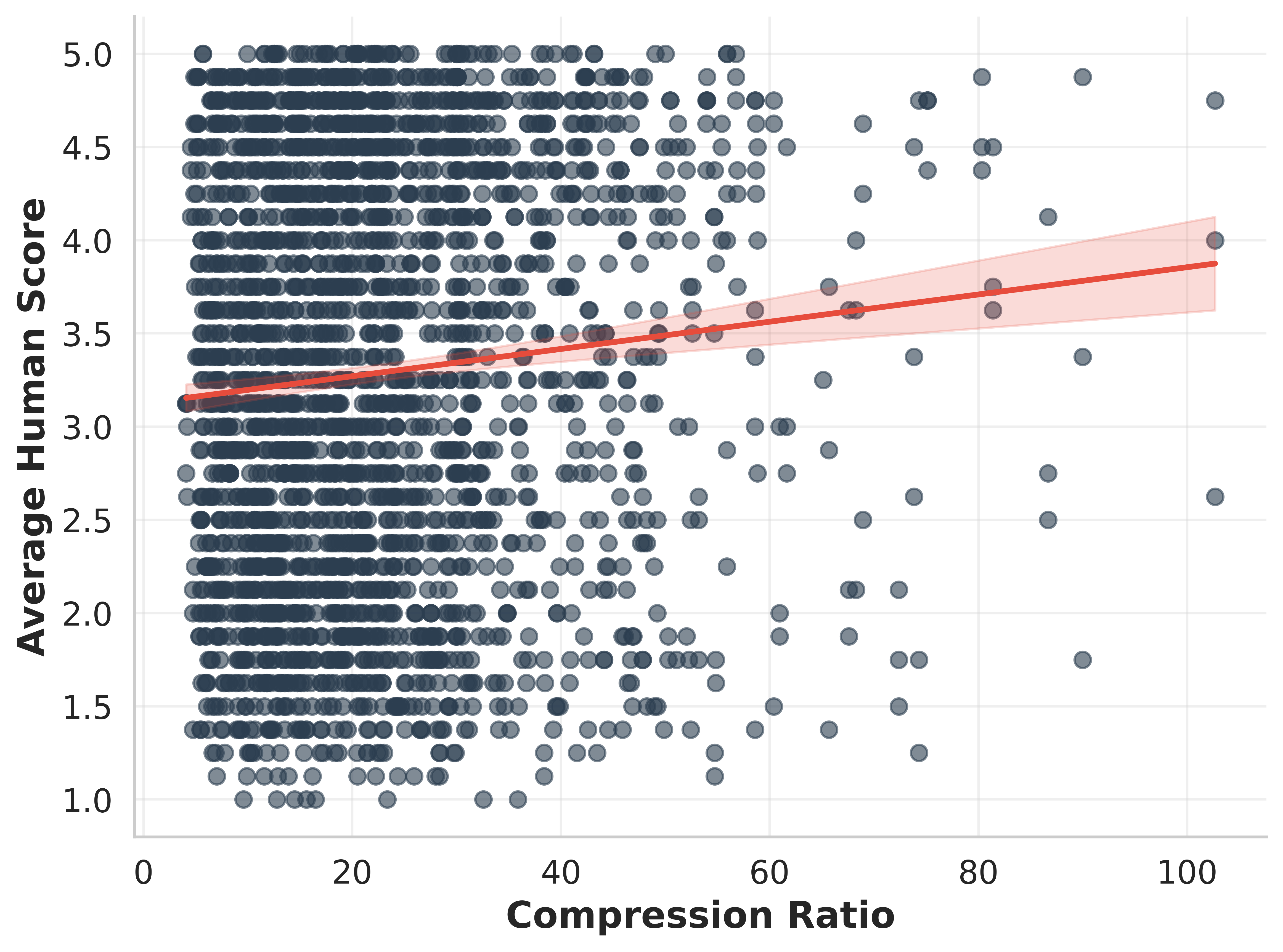}
  \vspace{-0.5cm}
  \caption{Impact of summary compression on human evaluation scores.}
  \vspace{-0.3cm}
  \label{fig:Compression_Ratio_Effect}
\end{figure}

Figure~\ref{fig:Compression_Ratio_Effect} represents the relationship between summary compression (ratio of article tokens to human summary tokens) and average human evaluation scores. Individual points represent observations, the red line shows the linear regression, and the shaded region indicates the 95\% confidence interval. The positive slope (0.007) imply a modest preference for summaries that are relatively more concise than their source texts.

\subsection{Entity Noise Variants}
\begin{figure}[ht!]
  \includegraphics[width=0.95\columnwidth]{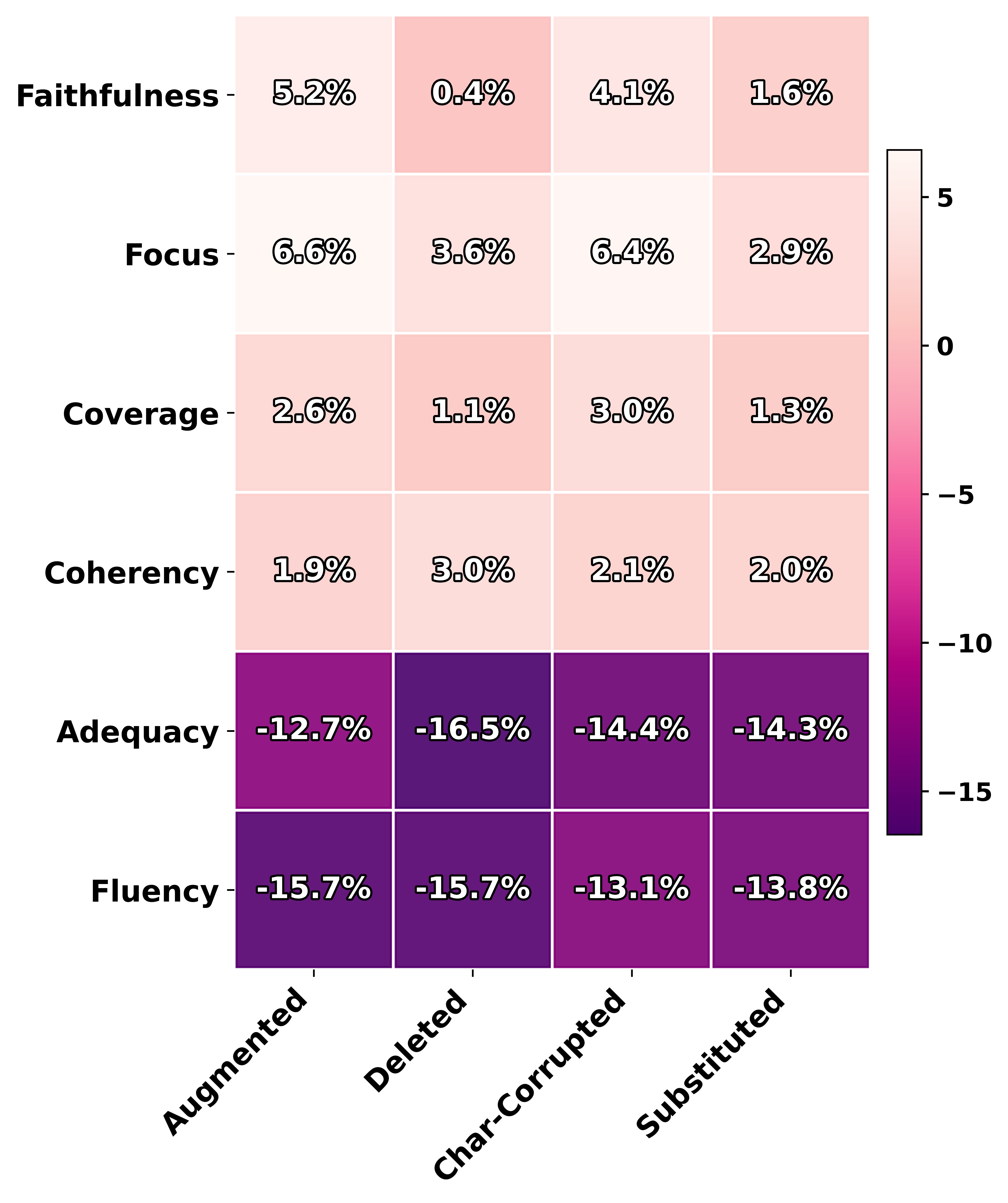}
  \vspace{-0.2cm}
  \caption{Relative shifts in human–COMET correlations across distinct entity noise conditions.}
  \vspace{-0.3cm}
  \label{fig:COMET_Corr_NER_Types}
\end{figure}

Figure~\ref{fig:COMET_Corr_NER_Types} shows the relative changes in Pearson correlation between human and COMET scores under different types of entity noise. In MT, entity perturbations substantially reduce correlations, with entity removal causing the steepest declines in both \textit{Adequacy} and \textit{Fluency}. In contrast, for TS, entity noise has a negligible impact, even yielding slight improvements in correlation.

\subsection{Additional Metric Sensitivity}
\label{sec:app_additional_metric_senitivity}

\begin{figure*}[t!]
  \includegraphics[width=\textwidth]{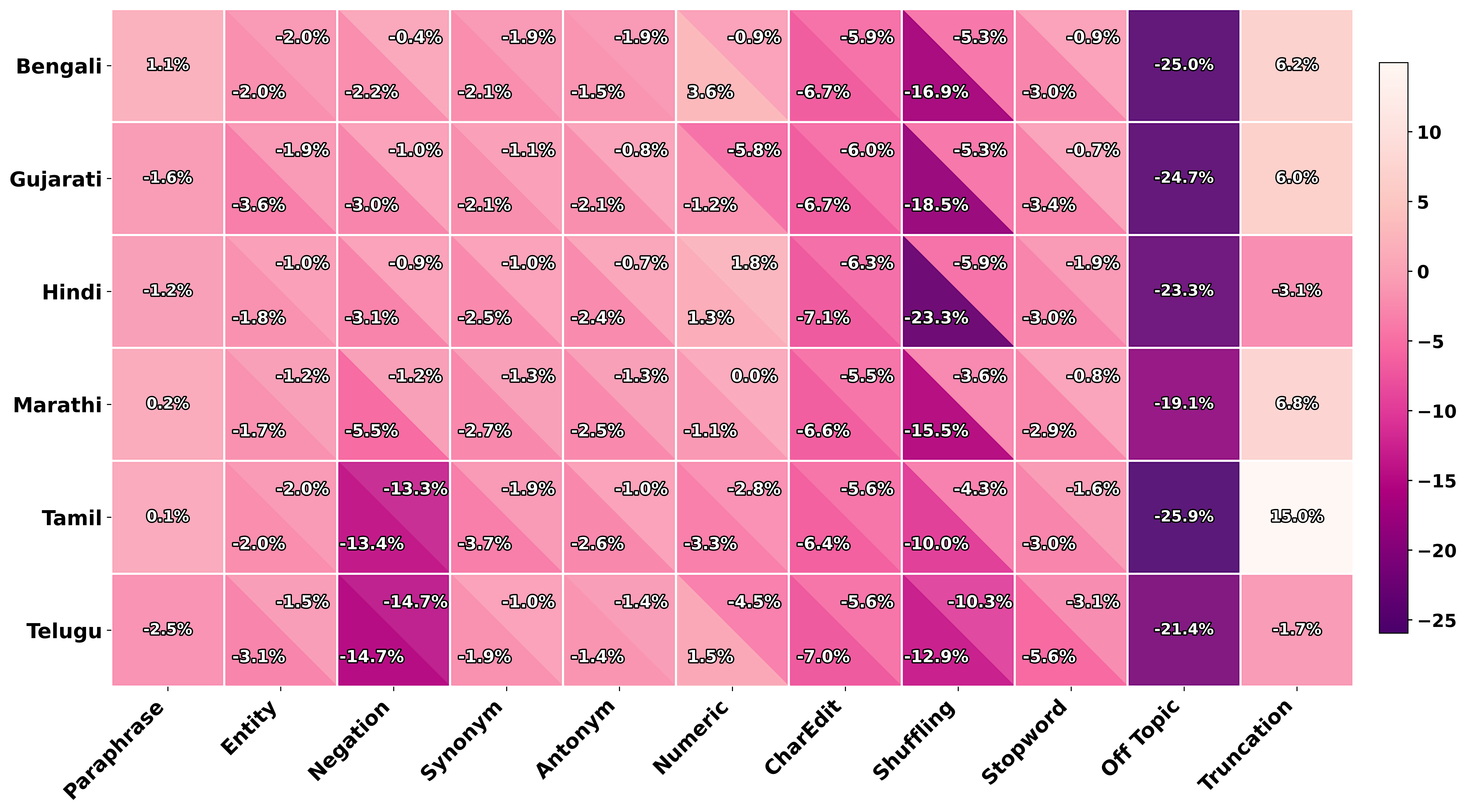}
  \vspace{-0.5cm}
  \caption{Relative ROUGE score drops (\%) across languages under diverse perturbations (top-right: TS; bottom-left: MT).}
  \vspace{-0.3cm}
  \label{fig:ROUGE_Drop_Lang}
\end{figure*}

Figure~\ref{fig:ROUGE_Drop_Lang} presents the relative ROUGE score drops across languages under various noise perturbations. In MT, shuffling induces the most severe degradation, whereas in TS, inserting off-topic sentences causes the largest decline. Similar to COMET, the effect of negation varies by language: Tamil and Telugu show marked sensitivity, while other languages remain comparatively robust. As anticipated, ROUGE is overall more susceptible to noise than COMET.

\begin{figure*}[h!]
  \includegraphics[width=\textwidth,keepaspectratio]{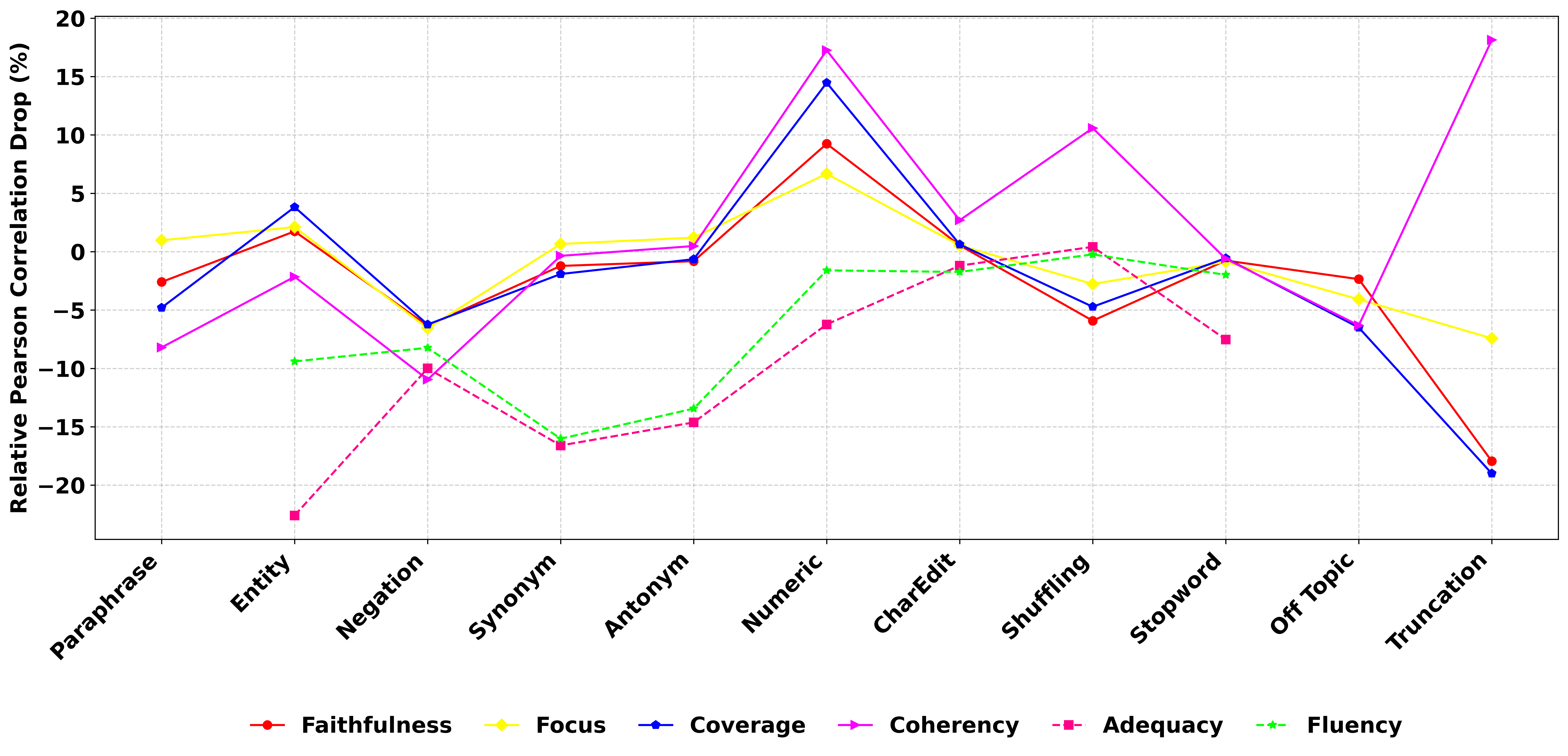}
  \vspace{-0.5cm}
  \caption{Relative Pearson correlation drop (\%) between human judgments and ROUGE across evaluation aspects under perturbations}
  \vspace{-0.3cm}
  \label{fig:ROUGE_Corr_Aspect}
\end{figure*}

Figure~\ref{fig:ROUGE_Corr_Aspect} illustrates the relative decreases in human-ROUGE correlations across different evaluation aspects. For MT, entity perturbations most strongly impact \textit{Adequacy}, whereas synonym replacements chiefly reduce \textit{Fluency}. In TS, consistent with COMET’s trend, truncating key sentences notably lowers correlations for \textit{Faithfulness}, \textit{Focus}, and \textit{Coverage}, while \textit{Coherency} unexpectedly improves. Overall, \textit{Focus} emerges as the most robust aspect, and \textit{Adequacy} as the least stable.

\begin{figure*}[ht!]
  \includegraphics[width=\textwidth]{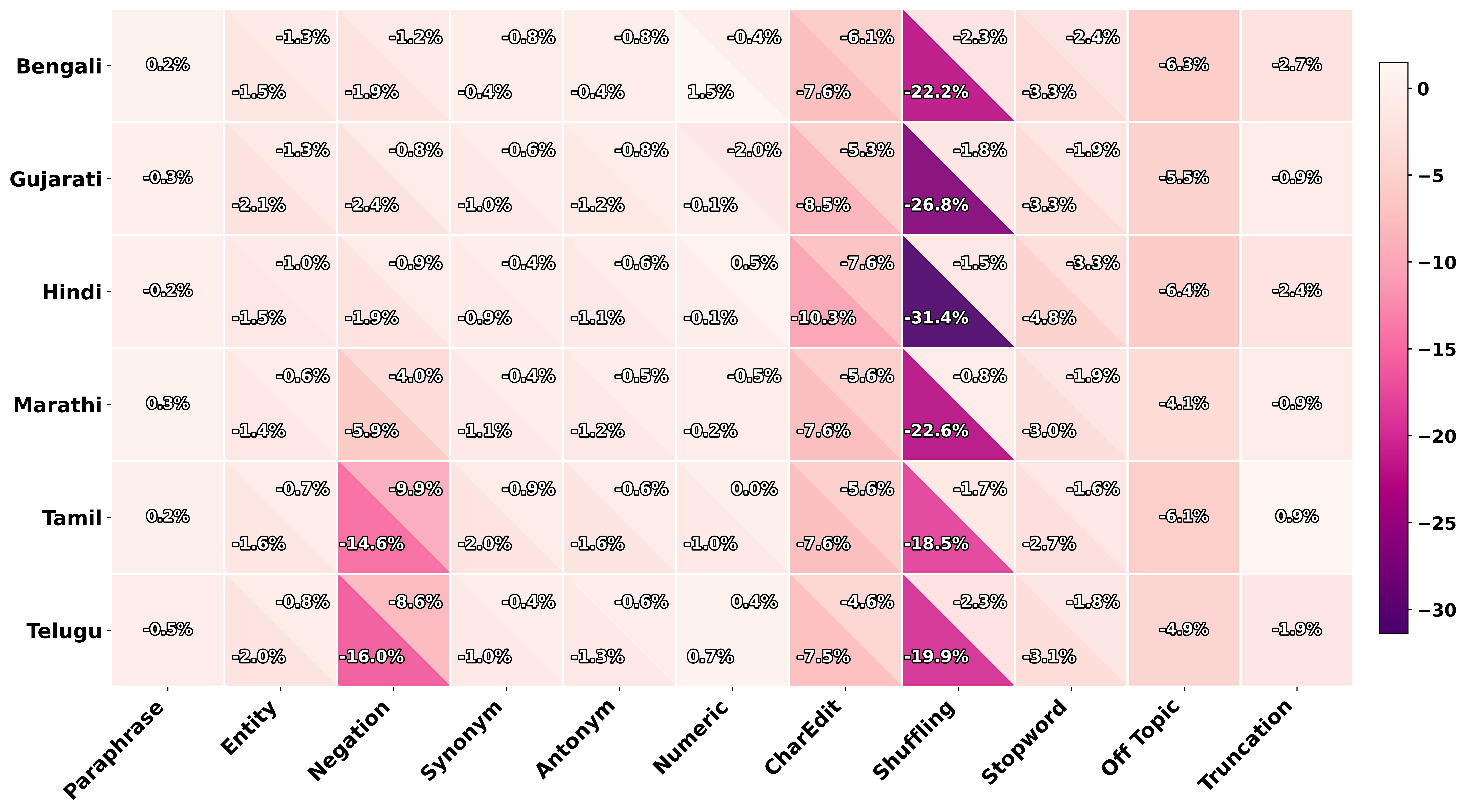}
  \vspace{-0.5cm}
  \caption{Relative BERTScore score drops (\%) across languages under diverse perturbations (top-right: TS; bottom-left: MT).}
  \vspace{-0.3cm}
  \label{fig:bertscore_Drop_Lang}
\end{figure*}

Figure~\ref{fig:bertscore_Drop_Lang} shows the relative BERTScore declines across languages under diverse perturbations. In MT, shuffling remains the most detrimental, while in TS, character-level edits cause the largest drop. Mirroring COMET’s behavior, negation effects are language-dependent: Tamil and Telugu again demonstrate heightened vulnerability. As expected, ROUGE is more noise-sensitive than BERTScore.

\begin{figure*}[h!]
  \includegraphics[width=\textwidth,keepaspectratio]{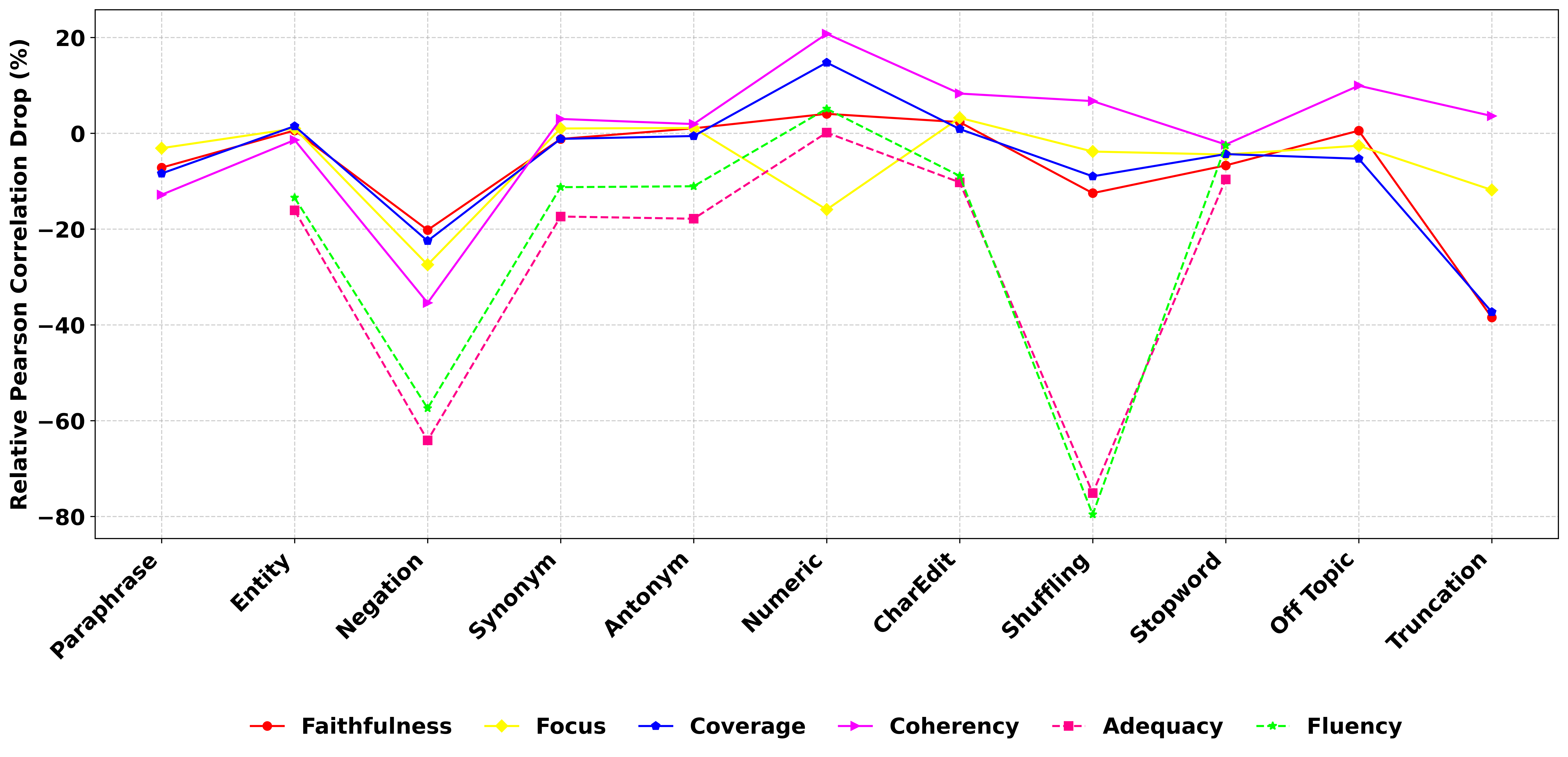}
  \vspace{-0.5cm}
  \caption{Relative Pearson correlation drop (\%) between human judgments and BERTScore across evaluation aspects under perturbations}
  \vspace{-0.3cm}
  \label{fig:bertscore_Corr_Aspect}
\end{figure*}

Figure~\ref{fig:bertscore_Corr_Aspect} highlights the decreases in human--BERTScore correlations by evaluation aspect. In MT, shuffling yields the sharpest reductions in both \textit{Adequacy} and \textit{Fluency}, whereas in TS, negation uniformly lowers all correlations. Collectively, paralleling ROUGE, \textit{Focus} proves most resilient, while \textit{Adequacy} remains the weakest dimension.

\end{document}